\documentclass[10pt,twocolumn,letterpaper]{article}

\usepackage{cvpr}      % To produce the REVIEW version
\usepackage{graphicx}
\usepackage{amsmath}
\usepackage{amssymb}
\usepackage{booktabs}

\usepackage[accsupp]{axessibility}  % Improves PDF readability for those with disabilities.

\usepackage[pagebackref,breaklinks,colorlinks]{hyperref}

\usepackage[capitalize]{cleveref}
\crefname{section}{Sec.}{Secs.}
\Crefname{section}{Section}{Sections}
\Crefname{table}{Table}{Tables}
\crefname{table}{Tab.}{Tabs.}

\usepackage{multirow}

\def\cvprPaperID{4276} % *** Enter the CVPR Paper ID here
\def\confName{CVPR}
\def\confYear{2023}

\begin{document}

\title{Task Difficulty Aware Parameter Allocation \& Regularization for Lifelong
Learning}

\author{Wenjin Wang, Yunqing Hu, Qianglong Chen, Yin Zhang\thanks{Corresponding author: Yin Zhang.}\\
Zhejiang University, Hangzhou, China\\
{\tt\small \{wangwenjin,yunqinghu,chenqianglong,zhangyin98\}@zju.edu.cn}
}
\maketitle

%%%%%%%%% ABSTRACT
\begin{abstract}
Parameter regularization or allocation methods are effective in overcoming catastrophic forgetting in lifelong learning.
However, they solve all tasks in a sequence uniformly and ignore the differences in the learning difficulty of different tasks.
So parameter regularization methods face significant forgetting when learning a new task very different from learned tasks, and parameter allocation methods face unnecessary parameter overhead when learning simple tasks.
In this paper, we propose the \emph{\textbf{P}arameter \textbf{A}llocation \& \textbf{R}egularization (PAR)}, which adaptively select an appropriate strategy for each task from parameter allocation and regularization based on its learning difficulty.
A task is easy for a model that has learned tasks related to it and vice versa.
We propose a divergence estimation method based on the Nearest-Prototype distance to measure the task relatedness using only features of the new task.
Moreover, we propose a time-efficient relatedness-aware sampling-based architecture search strategy to reduce the parameter overhead for allocation.
Experimental results on multiple benchmarks demonstrate that, compared with SOTAs, our method is scalable and significantly reduces the model's redundancy while improving the model's performance.
Further qualitative analysis indicates that PAR obtains reasonable task-relatedness.
\end{abstract}

%%%%%%%%% BODY TEXT
\section{Introduction}
\label{sec:intro}
\begin{figure}[t]
\small
\centering
\includegraphics[width=1\columnwidth]{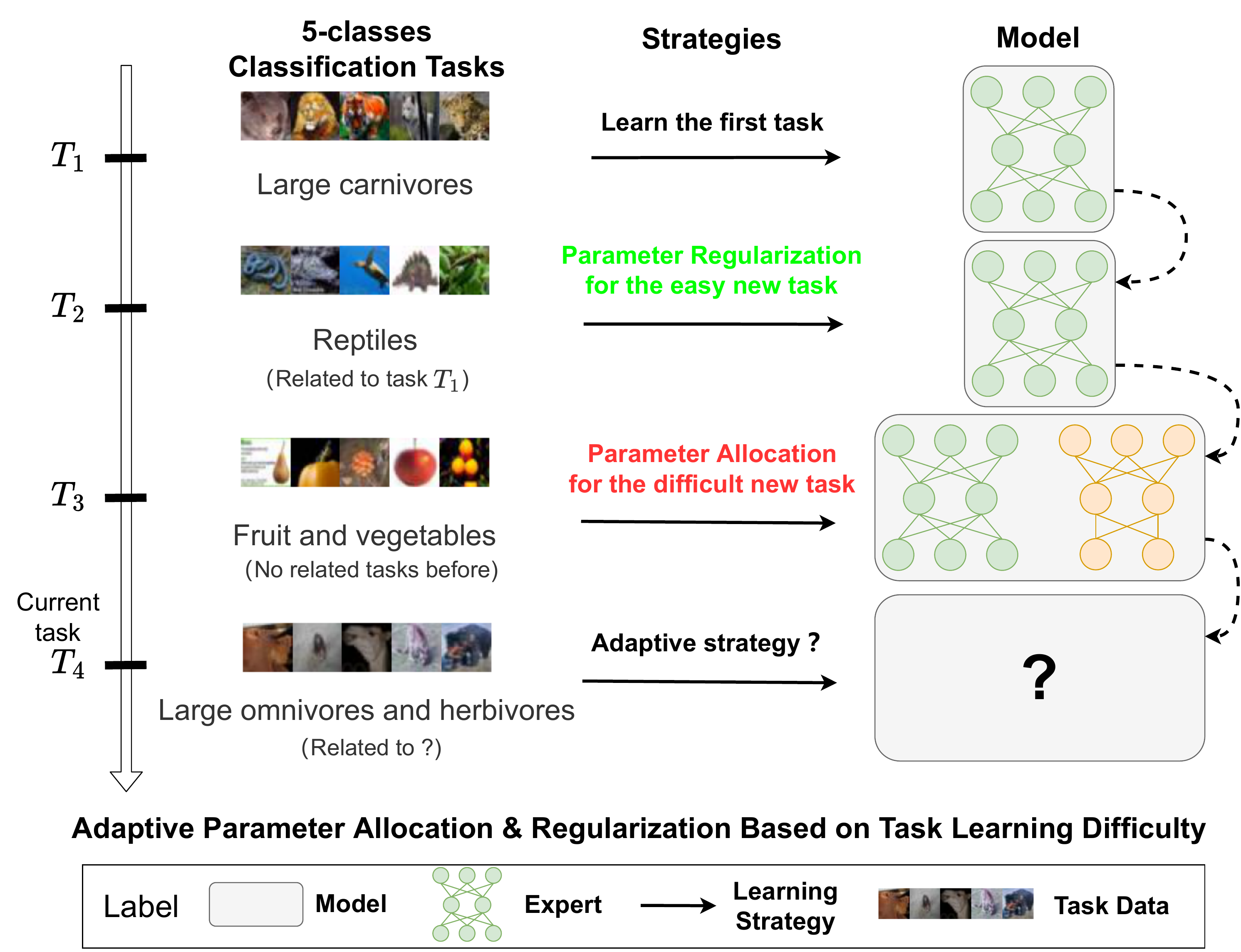}
\caption{
PAR adaptively selects a strategy from regularization and allocation to handle each task according to its learning difficulty.
The difficulty depends not only on the task itself, but also on whether the model has previously learned tasks related to it.
}
% \caption{In lifelong learning, the learning difficulty of each task is different and depends not only on itself but also on tasks the model has learned before.
% A natural idea is to allow the model adaptively select an appropriate strategy to handle each task according to its learning difficulty.
% For example, task $T_2$ (Reptiles) is easy for the model that has learned task $T_1$ (Large carnivores) and parameter regularization is sufficient to adapt to it.
% On the contrary, task $T_3$ (fruit and vegetables) is still difficult for the model that has learned tasks $T_1$ and $T_2$ and parameter allocation is necessary for it.}
\label{fig:parameter_allocation_regularization_idea}
\end{figure}
Recently, the lifelong learning \cite{delangeContinualLearningSurvey2021} ability of neural networks, i.e., learning continuously from a continuous sequence of tasks, has been extensively studied. It is natural for human beings to constantly learn and accumulate knowledge from tasks and then use it to facilitate future learning. However, classical models ~\cite{krizhevskyImageNetClassificationDeep2012,heDeepResidualLearning2016,vaswaniAttentionAllYou2017} suffer \emph{catastrophic forgetting}~\cite{frenchCatastrophicForgettingConnectionist1999}, i.e., the model's performance on learned tasks deteriorates rapidly after learning a new one. 

To overcome the catastrophic forgetting, many parameter regularization or allocation methods have been proposed.
\emph{Parameter regularization methods} \cite{kirkpatrickEWC2017, leeOvercomingCatastrophicForgetting2017a,liLearningForgetting2017,nguyenVARIATIONALCONTINUALLEARNING2018,dharLearningMemorizing2019, mengADINetAttributeDriven2020,liuOvercomingCatastrophicForgetting2020} alleviate forgetting by adding a regularization term to the loss function and perform well when the new task does not differ much from learned tasks.
\emph{Parameter allocation methods} based on static models \cite{serraOvercomingCatastrophicForgetting2018,rajasegaranRandomPathSelection2019,chenMitigatingForgettingOnline2020,keContinualLearningMixed2020} and dynamic models \cite{rusuProgressiveNeuralNetworks2016,aljundiExpertGateLifelong2017,liLearnGrowContinual2019,wenBatchEnsembleAlternativeApproach2019,veniatEfficientContinualLearning2020a,singhRectificationBasedKnowledgeRetention2021,singhCalibratingCNNsLifelong2020,qinBNSBuildingNetwork2021,yanDynamicallyExpandableRepresentation2021a,mendezLifelongLearningCompositional2020,miaoContinualLearningFilter2021,ostapenkoContinualLearningLocal2021} allocate different parameters to different tasks and can adapt to new tasks quite different from learned tasks. 
% However, they treat all tasks in a sequence equally and try to solve them uniformly, and ignore differences in the learning difficulty of different tasks in the sequence.
However, the above methods solve all tasks in a sequence uniformly, and ignore the differences of learning difficulty of different tasks. This leads to the significant forgetting in parameter regularization methods when learning a new task which is quite different from learned tasks, and also leads to unnecessary parameter cost in parameter allocation methods when learning some simple tasks.

In this paper, we propose a difficulty-aware method \textbf{P}arameter \textbf{A}llocation \& \textbf{R}egularization (PAR).
As shown in \cref{fig:parameter_allocation_regularization_idea}, we assume that the learning difficulty of a task in continual learning depends not only on the task itself, but also on the accumulated knowledge in the model.
A new task is easy to adapt for a model if it has learned related tasks before and vice versa.
Based on the assumption, the PAR adaptively adopts parameter allocation for difficult tasks and parameter regularization for easy tasks.
Specifically, the PAR divides tasks into \emph{task groups} and assigns each group a dedicated \emph{expert} model.
Given a new task, the PAR measures the relatedness between it and existing task groups at first.
If the new task is related to one of the existing groups, it is easy for the corresponding expert.
The PAR adds the task to the related group and learns it by the expert via the parameter regularization.
Otherwise, the new task is difficult for all existing experts, and the PAR assigns it to a new task group and allocates a new expert to learn it.

There are two challenges in this work: the measurement of relatedness and the parameter explosion associated with parameter allocation.
For the first one, we try to measure the relatedness by the KL divergence between feature distributions of tasks.
However, the KL divergence is intractable and needs to be estimated since the feature distributions of tasks are usually unknown.
In addition, the constraint of lifelong learning that only data of the current task are available exacerbates the difficulty of estimation.
To solve above problems, inspired by the divergence estimation based on k-NN distance \cite{wangDivergenceEstimationMultidimensional2009a}, we propose the divergence estimation method based on prototype distance, which only depends on the data of the current task.
For the second one, we try to reduce parameter overhead per expert by searching compact architecture for it.
However, the low time and memory efficiency is an obstacle to applying architecture search for a sequence of tasks in lifelong learning. To improve the efficiency of architecture search, we propose a relatedness-aware sampling-based hierarchical search.
The main contributions of this work are as follows:
\begin{itemize}
    \item We propose a lifelong learning framework named Parameter Allocation \& Regularization (PAR), which selects an appropriate strategy from parameter allocation and regularization for each task based on the learning difficulty.
    The difficulty depends on whether the model has learned related tasks before.
    \item We propose a divergence estimation method based on prototype distance to measure the distance between the new task and previous learned tasks with only data of the new task. 
    Meanwhile, we propose a relatedness-aware sampling-based architecture search to reduce the parameter overhead of parameter allocation.
    \item  Experimental results on CTrL, Mixed CIFAR100 and F-CelebA, CIFAR10-5, CIFAR100-10, CIFAR100-20 and MiniImageNet-20 demonstrate that PAR is scalable and significantly reduces the model redundancy while improving the model performance. Exhaustive ablation studies show the effectiveness of components in PAR and the visualizations show the reasonability of task distance in PAR.
\end{itemize}

\section{Related Work}
\label{sec:related_work}
\subsection{Lifelong Learning}
Many methods have been proposed to overcome catastrophic forgetting. 
\emph{Replay methods} try to replay samples of previous tasks when learning a new task from an \emph{episodic memory} \cite{rebuffiICaRLIncrementalClassifier2017,tangGraphBasedContinualLearning2020} or a \emph{generative memory} \cite{ostapenkoLearningRememberSynaptic2019,shinContinualLearningDeep2017,ayubEECLearningEncode2020}.
\emph{Parameter regularization methods}, including the \emph{prior-focused regularization} \cite{kirkpatrickEWC2017,leeOvercomingCatastrophicForgetting2017a,nguyenVARIATIONALCONTINUALLEARNING2018,liuOvercomingCatastrophicForgetting2020} and the \emph{data-focused regularization} \cite{liLearningForgetting2017,dharLearningMemorizing2019,mengADINetAttributeDriven2020},
try to alleviate forgetting by introducing a regularization term in the loss function of the new task.
\emph{Parameter allocation methods} based on the \emph{static model} \cite{serraOvercomingCatastrophicForgetting2018,rajasegaranRandomPathSelection2019,chenMitigatingForgettingOnline2020,keContinualLearningMixed2020} and the \emph{dynamic model} \cite{rusuProgressiveNeuralNetworks2016,aljundiExpertGateLifelong2017,liLearnGrowContinual2019,wenBatchEnsembleAlternativeApproach2019,veniatEfficientContinualLearning2020a,singhRectificationBasedKnowledgeRetention2021,singhCalibratingCNNsLifelong2020,qinBNSBuildingNetwork2021,yanDynamicallyExpandableRepresentation2021a,mendezLifelongLearningCompositional2020,miaoContinualLearningFilter2021,ostapenkoContinualLearningLocal2021} overcome catastrophic forgetting by allocating different parameters to different tasks.

% \paragraph{Methods with relatedness}
\textbf{Methods with relatedness.}
Several methods also consider the utility of task relatedness \cite{aljundiExpertGateLifelong2017,keContinualLearningMixed2020}.
Expert Gate \cite{aljundiExpertGateLifelong2017} assigns dedicated experts and auto-encoders for tasks and calculates the task relatedness by reconstruction error of auto-encoders. The task relatedness is used to transfer knowledge from the most related previous task.
CAT \cite{keContinualLearningMixed2020} defines the task similarity by the positive knowledge transfer.
It focuses on selectively transferring the knowledge from similar previous tasks and dealing with forgetting between dissimilar tasks by hard attention.
In this paper, we propose a divergence estimation method based on prototype distance to calculate the task relatedness used to measure the task learning difficulty.

% \subsubsection{Methods with NAS}
% \paragraph{Methods with NAS}
\textbf{Methods with NAS.}
Static model based methods \cite{rajasegaranRandomPathSelection2019,chenMitigatingForgettingOnline2020} try to search a sub-model for each task with neural architecture search (NAS).
Dynamic model based methods \cite{xuReinforcedContinualLearning2018,liLearnGrowContinual2019,veniatEfficientContinualLearning2020a} adopt NAS to select an appropriate model expansion strategy for each task and face the high GPU memory, parameter, and time overhead.
Instead, in this paper, we propose a relatedness-aware hierarchical cell-based architecture search to search compact architecture for each expert with a lower GPU memory and time overhead.
\cite{cheng2018learning,huang2018gnas} adopt NAS on multi-task learning.

\subsection{Cell-based NAS}
\label{sec:related_work_nas}
Neural architecture search (NAS) \cite{elskenNeuralArchitectureSearch2019,renComprehensiveSurveyNeural2020} aims to search for efficient neural network architectures from a pre-defined search space in a data-driven manner.
To reduce the size of the search space, \emph{cell-based} NAS methods \cite{zophLearningTransferableArchitectures2018,liuDARTSDIFFERENTIABLEARCHITECTURE2018,zhengMultinomialDistributionLearning2019} try to search for a cell architecture from a pre-defined cell search space, where a cell is a tiny convolutional network mapping an $H\times W\times F$ tensor to another $H'\times W'\times F'$. 
The final model consists of a pre-defined number of stacked cells.
The cell in NAS is similar to the residual block in the residual network (ResNet), but its architecture is more complex and is a directed acyclic graph (DAG). 
The operations in the search space are usually parameter-efficient.
NAS methods usually produce more compact architectures than hand-crafted designs.
% In this paper, we adopt cell-based NAS to search compact cell architectures for experts, which reduces the parameter overhead in the parameter allocation phase. 

\label{sec:method}
\begin{figure*}[t]
\centering
% \small
\includegraphics[width=0.9\textwidth]{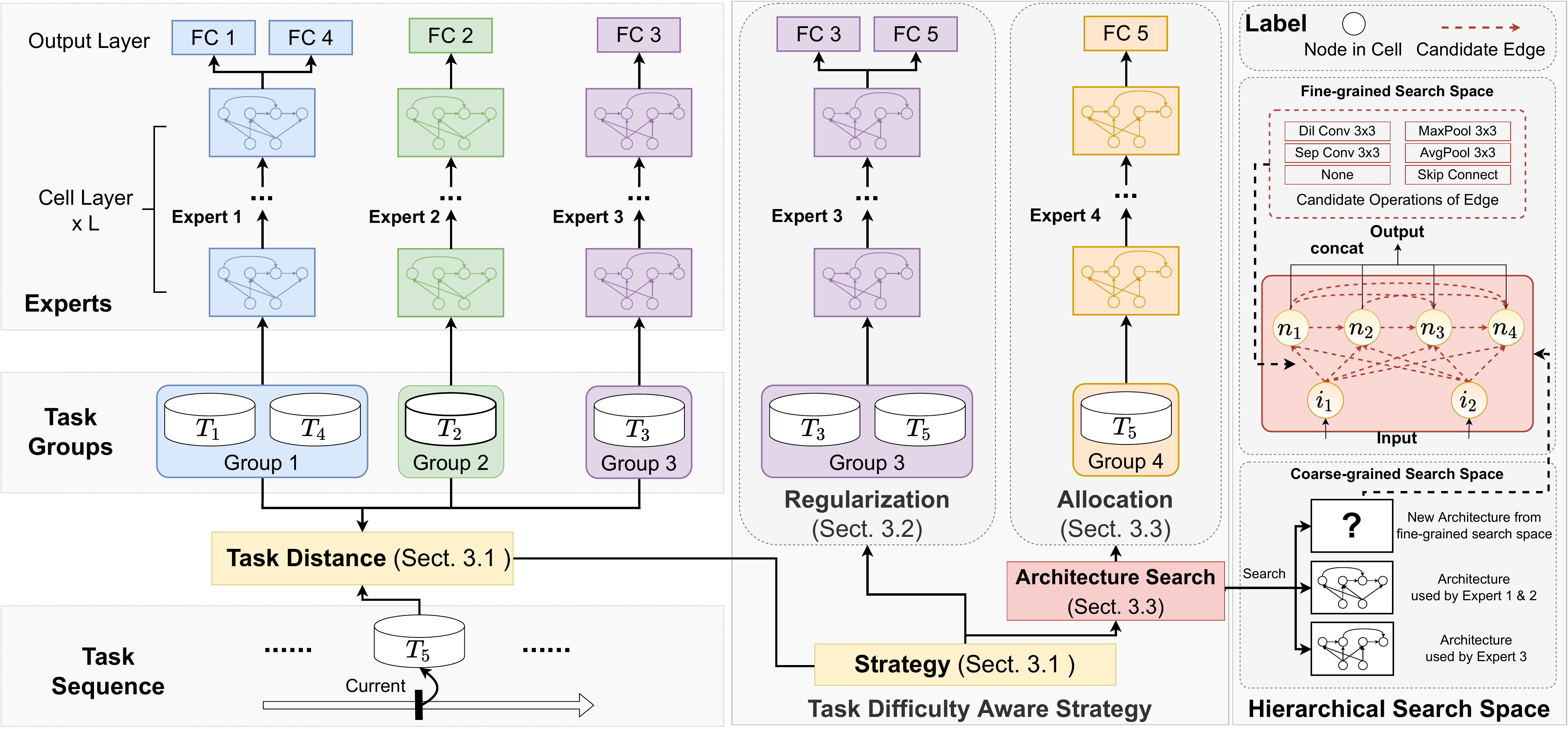}
\caption{\textbf{The architecture of PAR.}
PAR divides previously learned tasks into task groups and each group has a dedicated expert model.
Given a new task $T_5$, PAR calculates the distance between it and existing task groups to measure its learning difficulty. Then, PAR learns the new task by parameter regularization if it is easy to learn. Otherwise, PAR learns the new task by parameter allocation and search compact architecture for each expert by a relatedness-aware hierarchical architecture search.}
\label{fig:parameter_allocation_regularization_architecture}
\end{figure*}

\section{Method}
We focus on the task-incremental scenario of lifelong learning. 
Specifically, the model learns a sequence of tasks $\mathcal{T}=\{T_1, \dots, T_t, \dots, T_N\}$ one by one and the task id of sample is available during both training and inference.
Each task $T_t$ has a training dataset, $D_{train}^t=\{(x_i^t,y_i^t);i=1,\dots,n_{train}^t\}$, where $y_i^t$ is the true label and $n_{train}^t$ is the number of training examples.
Similarly, we denote the validation and test dataset of task $T_t$ as $D_{valid}^t$ and $D_{test}^t$.

As shown in \cref{fig:parameter_allocation_regularization_architecture}, the PAR divides previously learned tasks into task groups $\mathcal{G}$ and assigns each group $\mathcal{G}_i$ a dedicated expert model $E_i$.
For a new task, PAR calculates the distances, which reflect the relatedness, between it and existing task groups at first.
Then, it adopt an appropriate learning strategy from parameter allocation and regularization for the new task based on the distances.

\subsection{Task Distance via Nearest-Prototype Distance}
\label{sec:method_task_relevance}
The PAR calculates the distance between the new task and each existing task at first, then it averages the distances of tasks in the same group as the distance between the new task and the group.

We calculate the distance between two tasks by the KL divergence between their distributions.
However, the KL divergence is intractable because the true distributions of tasks are usually agnostic.
Instead, we estimate the KL divergence by the features of data.
To enhance estimation stability, we introduce an extra pre-trained extractor\footnote{We adopt a ResNet18 pre-trained on the ImageNet. It does not introduce unfairness because the generated features are not involved in model training and prediction.} to generate a robust feature $X_i^t$ for each image $x_i^t$ in task $T_t$.
\emph{The pre-trained extractor is only used for divergence estimation and does not affect the learning of tasks.}
The number of extra parameters introduced by the model is a constant and does not affect the scalability of our method.

\paragraph{$k$-Nearest Neighbor Distance}
% \subsubsection{$k$-Nearest Neighbor Distance}
Given the features of distributions, a classical divergence estimation method for multidimensional densities is the $k$-Nearest Neighbor ($k$-NN) distance~\cite{wangDivergenceEstimationMultidimensional2009a}.
Suppose the feature distributions of the new task $T_i$ and an existing task $T_j$ are $q_i$ and $q_j$ respectively, and the features $X^i=\{X_1^i,\dots,X_n^i\}$ and $X^j=\{X_1^j,\dots,X_m^j\}$ are drawn independently from distributions $q_i$ and $q_j$, where $n$ and $m$ are the numbers of samples.
The $k$-NN distance of feature $X^i_l$ in $X^i$ and $X^j$ are denoted by $\nu_k(l)$ and $\rho_k(l)$ respectively. Specifically, the $\rho_k(l)$ is the Euclidean distance from $X^i_l$ to its $k$-NN in $\{X^i_z\}_{z\neq l}$. Similarly, the $\nu_k(l)$ is the distance from $X^i_l$ to its $k$-NN in $X^j$.
Then, the KL divergence between distributions $q_i$ and $q_j$ are estimated as follows:
\begin{equation}
    \label{eq:knn_kl}
    KL(q_i||q_j)\approx\widehat{KL}(q_i||q_j)=\frac{d}{n}\sum_{l=1}^n\log\frac{\nu_k(l)}{\rho_k(l)}+\log\frac{m}{n-1},
\end{equation}
where $d$ is the feature dimension.

The $k$-NN distance is asymmetric and this is consistent with the asymmetry of the KL divergence.
The motivation behind \cref{eq:knn_kl} is that the $k$-NN distance is the radius of a $d$-dimensional open Euclidean ball used for $k$-NN density estimation.
Readers can refer to \cite{wangDivergenceEstimationMultidimensional2009a} for more details such as convergence analysis.

% \subsubsection{Nearest-Prototype Distance}
\paragraph{Nearest-Prototype Distance}
However, the calculation of the $k$-NN distance involves features of two tasks $T_i$ and $T_j$ and violates the constraint of lifelong learning that only data of the current task $T_i$ are available.
To overcome this problem, we propose the Nearest-Prototype distance to replace the $k$-NN distance.
Specifically, suppose the set of classes in task $T^i$ is $C^i$, for each class $c\in C^i$, we maintain a prototype feature $U^i_c$, which is the mean of features of samples belonging to the class $c$.
Then, the Nearest-Prototype distance of $X^i_l$ to $X^i$ is defined as follows:
\begin{equation}
    \label{eq:np_distance_rho}
    \rho(l)=||X_l^i-U^i_{c(l)}||,
\end{equation}
where $||\cdot||$ is the Euclidean distance and $c(l)$ is the class of $X_l^i$.
Similarly, we denote the set of classes in task $T_j$ by $C^j$ and the prototype features of $X^j$ for each class $c\in C^j$ by $U_c^j$.
Then, the Nearest-Prototype distance of $X_l^i$ to $X^j$ is denoted as
\begin{equation}
    \label{eq:np_distance_nu}
    \nu(l)=\min_c||X^i_l-U^j_c||_2,c\in C^j.
\end{equation}

Then, by replacing the $k$-NN distances $\nu_k(l)$ and $\rho_k(l)$ in \cref{eq:knn_kl} by the the Nearest-Prototype distances $\nu(l)$ and $\rho(l)$, the KL divergence estimated by the Nearest-Prototype distance is as follows:
\begin{equation}
    \label{eq:np_kl}
    KL(q_i||q_j)\approx\widetilde{KL}(q_i||q_j)=\sum_{c\in C^i}\frac{1}{n_c}\sum_{l=1}^{n_c}\log\frac{\nu(l)}{\rho(l)},
\end{equation}
where constant terms are omitted because we are only concerned with the relative magnitude of KL divergence.
The \cref{eq:np_distance_rho,eq:np_distance_nu,eq:np_kl} only involve features of the new task $T_i$ and prototypes of the existing task $T_j$, which satisfies the constraint of lifelong learning.
The storage cost of prototypes is small and negligible.

The motivation behind \cref{eq:np_kl} is intuitive.
The Nearest-Prototype distance $\rho(l)$ reflects the relationship between the sample $X_l^i$ in task $T_i$ and its class prototype in feature space.
The Nearest-Prototype distance $\nu(l)$ reflects the relationship between the sample $X_l^i$ and class prototypes of existing task $T_j$ in feature space.
If the value of $\rho(l)$ and $\nu(l)$ are equal for each $X_l^i$, the class prototypes of two tasks are close.
In this case, the distribution of the two tasks are similar, and the estimated KL divergence by \cref{eq:np_kl} is $0$.

\paragraph{Adaptive Learning Strategy}
% \subsubsection{Adaptive Learning Strategy}
Given the distance between the new task $T_i$ and each existing tasks by KL divergence, we denote the distance $s'_{i,g}$ between the task $T_i$ and a task group $\mathcal{G}_g$ by the average of distances of tasks in the group as follows:
\begin{equation}
    \label{eq:relevance_task_group}
    s'_{i,g}=\frac{1}{|\mathcal{G}_g|}\sum_{j'}^{|\mathcal{G}_g|}KL(q_i||q_{j'}),
\end{equation}
where $q_i$ and $q_{j'}$ represent feature distributions of task $T_i$ and the $j'$-th task in group $\mathcal{G}_g$.
The $s'_{i,g}\in [0, \infty)$ since the range of the KL divergence is $[0, \infty)$.
However, we try to use the distance to reflect the relatedness, which is usually measured by a value between $0$ and $1$.
So we map the $s'_{i,g}$ into $[0,1]$ by a monotone increasing function as follows:
\begin{equation}
    \label{eq:map}
    s_{i,g}=\min(s'_{i,g}, 1-e^{-2\times s'_{i,g}}),
\end{equation}
where $e$ is the Euler's number.

Suppose the smallest distance between the new task $T_t$ and existing groups is $s_{t,g^*}$ and the corresponding task group is $\mathcal{G}_{g*}$, the PAR selects the learning strategy by comparing the $s_{t,g^*}$ with a hyper-parameter $\alpha$.
If $s_{t,g^*}\leq\alpha$, the new task is added to this related group $\mathcal{G}_{g^*}$ and learned by parameter regularization.
Otherwise, the new task is assigned to a new group and learned by parameter allocation because no existing task group is related to it.

\subsection{Parameter Regularization}
\label{sec:method_parameter_regularization}
% tab:exp_ctrl
\begin{table*}[t]
\centering
\small
\begin{tabular}{@{}clcccccccccccc@{}}
\toprule
  &\multirow{2}{*}{Method} &
  \multicolumn{2}{c}{$\mathcal{S}^{-}$} &
  \multicolumn{2}{c}{$\mathcal{S}^{+}$} &
  \multicolumn{2}{c}{$\mathcal{S}^{\text{in}}$} &
  \multicolumn{2}{c}{$\mathcal{S}^{\text{out}}$} &
  \multicolumn{2}{c}{$\mathcal{S}^{\text{pl}}$} &
  \multicolumn{2}{c}{$\mathcal{S}^{\text{long}}$} \\
  &  & AP     & AF      & AP     & AF      & AP     & AF      & AP     & AF      & AP     & AF      & AP     & AF      \\
\midrule
&Finetune    & 0.180 & -0.330 & 0.240 & -0.230 & 0.180 & -0.310 & 0.150 & -0.370 & 0.210 & -0.300 & 0.200 & -0.400 \\
\midrule
(1)&EWC& 0.540 & -0.010 & 0.370 & -0.040 & 0.430 & -0.120 & 0.510 & -0.030 & 0.300 & -0.170 & 0.270 & -0.300 \\
\midrule
(2)&ER& 0.410 & -0.130 & 0.430 & -0.070 & 0.380 & -0.170 & 0.540 & -0.070 & 0.450 & -0.080 & -     & -      \\
\midrule
(3)&HAT$^*$& 0.570 & -0.010 & 0.570 & 0.000  & 0.580 & -0.010 & 0.600 & 0.000  & 0.580 & 0.000  & 0.240 & -0.100 \\
\midrule
\multirow{6}{*}{(4)}&Independent & 0.560 & 0.000  & 0.570 & 0.000  & 0.570 & 0.000  & 0.610 & 0.000  & 0.590 & 0.000  & 0.570 & 0.000  \\
&PNN& 0.620 & 0.000  & 0.520 & 0.000  & 0.570 & 0.000  & 0.620 & 0.000  & 0.540 & 0.000  & -     & -      \\
&SG-F& 0.636 & 0.000  & 0.615 & 0.000  & 0.655 & 0.000  & 0.641 & 0.000  & 0.620 & 0.000  & -     & -      \\
&MNTDP-S& 0.630 & 0.000  & 0.560 & 0.000  & 0.570 & 0.000  & 0.640 & 0.000  & 0.550 & 0.000  & 0.680 & 0.000  \\
&MNTDP-D& 0.670 & 0.000  & 0.610 & 0.000  & 0.600 & 0.000  & 0.680 & 0.000  & 0.620 & 0.000  & 0.750 & 0.000  \\
&LMC& 0.666 & 0.000  & 0.601 & -0.014 & 0.695 & 0.000  & 0.667 & -0.010 & 0.616 & -0.035 & 0.639 & -      \\
\midrule
&PAR(our)    & \textbf{0.706} & 0.000  & \textbf{0.670} & -0.017  & \textbf{0.699} & 0.000  & \textbf{0.694} & 0.000  & \textbf{0.700} & 0.000  &  \textbf{0.773} & -0.016 \\
\bottomrule
\end{tabular}
\caption{
The average performance (AP) and average forgetting (AF) on the six streams ($\mathcal{S}^{-}$, $\mathcal{S}^{+}$, $\mathcal{S}^{\text{in}}$, $\mathcal{S}^{\text{out}}$, $\mathcal{S}^{\text{pl}}$, $\mathcal{S}^{\text{long}}$) in the CTRL benchmark.
We compare the PAR with four types of baselines: (1) parameter regularization, (2) replay, (3) parameter allocation with static model, and (4) parameter allocation with dynamic model.
The * corresponds to model using the AlexNet backbone.}
\label{tab:exp_ctrl}
\end{table*}
A new task $T_t$ is easy for the expert model $E_{g^*}$ if it is related to the group $G_{g^*}$. 
PAR reuses the expert $E_{g^*}$ to learn this task by parameter regularization.
Inspired by LwF \cite{liLearningForgetting2017}, we adopt a parameter regularization method based on knowledge distillation.
Specifically, the loss function consists of a training loss $\mathcal{L}_\text{new}$ and a distillation loss $\mathcal{L}_\text{old}$.
The training loss encourages expert $E_{g^*}$ to adapt to the new task $T_t$ and is the cross-entropy for classification as follows:
\begin{equation}
    \label{eq:loss_pr_new}
    \mathcal{L}_\text{new}=-\sum_{(x,y)\in D_{train}^t}\log(p_{E_{g^*}}(y|x)).
\end{equation}
The distillation loss encourages the expert $E_{g^*}$ to maintain performance on previous tasks in the group $G_{g^*}$.
To calculate it, we record the logits $\mathbf{y}^j$ of the output head of each previous task $T_j$ for each sample $x$ of task $T_t$.
The distillation loss is as follows:
\begin{equation}
    \label{eq:loss_pr_dl}
    \mathcal{L}_\text{old}=-\sum_{(x,y)\in D_{train}^t}\sum_{T_j\in \mathcal{G}_{g^*}}\mathbf{y}^j\cdot\log(\mathbf{p_{E_{g^*}}(x)}^j),
\end{equation}
where $\mathbf{y}^j$ and $\mathbf{p_{E_{g^*}}(x)}^j$ are vectors with lengths equal to the number of categories of the previous task $T_j$.
The total loss is 
\begin{equation}
    \label{eq:loss_pr}
    \mathcal{L}_\text{PR}=\mathcal{L}_\text{new}+\lambda\mathcal{L}_\text{old},
\end{equation}
where $\lambda$ is a hyper-parameter to balance training and distillation loss.
\emph{Note that our method is without memory and does not require storing samples for previous tasks.}

However, the expert $E_{g^*}$ may over-fit the new task whose sample size is far less than tasks in the group $\mathcal{G}_{g^*}$, even though the new task is related to the group.
This leads to interference with previously accumulated knowledge in the expert $E_{g^*}$.
To avoid the above problem, PAR records the maximum sample size of tasks in each task group.
Suppose the maximum sample size in the group $\mathcal{G}_{g^*}$ is $Q$, PAR freezes the parameters of expert $E_{g^*}$ except for the task-specific classification head during the learning if the sample size of the new task is less than 10 percent of $Q$.
By transferring the existing knowledge in expert $E_{g^*}$, only optimizing the classification header is sufficient to adapt to the new task because the new task is related to the group $\mathcal{G}_{g^*}$.

\subsection{Parameter Allocation}
\label{sec:method_parameter_allocation}
If no existing groups are related to the new task $T_t$, PAR assigns it to a new task group $\mathcal{G}_{M+1}$ with a new expert $E_{M+1}$, where $M$ is the number of existing groups.
We adopt the cross-entropy loss for the task $T_t$ as follows: 
\begin{equation}
    \label{eq:loss_pa}
    \mathcal{L}_\text{PA}=-\sum_{(x,y)\in D_{train}^t}\log(p_{E_{M+1}}(y|x)).
\end{equation}

The number of experts and parameters in PAR is proportional to the number of task groups, mitigating the growth of parameters.
To further reduce the parameter overhead of each expert, we adopt cell-based NAS (refer to \cref{sec:related_work_nas} for details) to search for a compact architecture for it.

Each expert in PAR is stacked with multiple cells and the search for expert architecture is equivalent to the search for the cell architecture.
Since the time overhead of NAS becomes unbearable as the number of tasks increases, we propose a relatedness-aware sampling-based architecture search strategy to improve efficiency.

As shown in \cref{fig:parameter_allocation_regularization_architecture}, we construct a hierarchical search space.
The coarse-grained search space contains cells used by existing experts and an unknown cell whose architecture will come from the fine-grained search space.
Following common practice \cite{liuDARTSDIFFERENTIABLEARCHITECTURE2018,zhengMultinomialDistributionLearning2019}, the fine-grained search space is a directed acyclic graph (DAG) with seven nodes (two input nodes $i_1,i_2$, an ordered sequence of intermediate nodes $n_1,n_2,n_3,n_4$, and an output node).
The input nodes represent the outputs of the previous two layers.
The output is concatenated from all intermediate nodes. 
Each intermediate node has a directed candidate edge from each of its predecessors. Each candidate edge is associated with six parameter-efficient candidate operations.

To search a cell for the new task, we introduce a hyper-parameter $\beta$. 
If $s_{t,g^*}\leq\beta$, PAR reuses the cell of expert $E_{g^*}$ for the task $T_t$.
A task distance greater than $\alpha$ and less than $\beta$ indicates that the new task is not related enough to share the same expert with tasks in the group $\mathcal{G}_{g^*}$ but can use the same architecture.
If $s_{t,g^*}>\beta$, PAR assigns the unknown cell to the new expert and adopts a sampling-based NAS method named MDL \cite{zhengMultinomialDistributionLearning2019} to determine its architecture.
Due to limited space, we leave the details of the MDL in the supplementary.

\section{Experiments}
\label{sec:experiment}
%tab:exp_cifar100_celeba
\begin{table}[t]
\centering
% \small
\begin{tabular}{cccccc}
\toprule
Method& Finetune& HAT& CAT& PAR\\
\midrule
AP& 0.6155& 0.6178& 0.6831& \textbf{0.7122}\\
\bottomrule
\end{tabular}
\caption{We compare the average performance (AP) of PAR with two parameter allocation methods based on static model on the mixed CIFAR100 and F-CelebA benchmark.}
\label{tab:exp_cifar100_celeba}
\end{table}

%tab:exp_cifar100-10_cifar10
\begin{table}[t]
\centering
\small
\begin{tabular}{cccccc}
\toprule
\multirow{2}{*}{\#}&
\multirow{2}{*}{Method}& \multicolumn{2}{c}{CIFAR100-10}&
\multicolumn{2}{c}{MiniImageNet-20}
\\
&
&AP(\%)
&AF(\%)
&AP(\%)
&AF(\%)
\\
\midrule
&Finetune& 0.280& -0.560& 0.378& -0.314\\
\midrule
\multirow{3}{*}{1}
&EWC
&0.608
&-0.061
&0.574
&-0.144
\\
&MAS
&0.579
&-0.050
&0.368
&-0.238
\\
&LwF
&0.655
&-0.088
&0.589
&-0.169
\\
\midrule
\multirow{2}{*}{2}
&GPM$^*$
&0.725
&-
&0.604
&-
\\
&
A-GEM&
0.677&
-0.120&
0.524&
-0.152\\
\midrule
\multirow{5}{*}{3}&
Independent&
0.820&
0.000&
0.787&
0.000\\
&PN&
0.821&
0.000&
0.781&
0.000\\
&Learn to Grow&
0.791&
0.000&
-&
-\\
&MNTDP-S& 0.750& -0.000& -& -\\
&MNTDP-D& 0.830& -0.000& -& -\\
\midrule
&PAR
&\textbf{0.853}
&-0.020
&\textbf{0.816}
&-0.022
\\
\bottomrule
\end{tabular}
\caption{
PAR outperforms (1) parameter regularization methods, (2) replay methods, and (3) parameter allocation methods with dynamic model on classical benchmarks. The * corresponds to model using the AlexNet backbone.}
\label{tab:exp_cifar100-10_cifar10}
\end{table}

\subsection{Experimental Settings}
% \subsubsection{Benchmarks}
\paragraph{Benchmarks}
We adopt two benchmarks the CTrL \cite{veniatEfficientContinualLearning2020a} and the Mixed CIFAR100 and F-CelebA \cite{keContinualLearningMixed2020}, which contain mixed similar and dissimilar tasks.
CTrL \cite{veniatEfficientContinualLearning2020a} includes 6 streams of image classification tasks: $S^-$ is used to evaluate the ability of direct transfer, $S^+$ is used to evaluate the ability of knowledge update, $S^\text{in}$ and $S^\text{out}$ are used to evaluate the transfer to similar input and output distributions respectively, $S^\text{pl}$ is used to evaluate the plasticity, $S^\text{long}$ consists of 100 tasks and is used to evaluate the scalability.
Similarly, Mixed CIFAR100 and F-CelebA \cite{keContinualLearningMixed2020} including mixed 10 similar tasks from F-CelebA and 10 dissimilar tasks from CIFAR100 \cite{krizhevskyLearningMultipleLayers2009}.

Further, we adopt classical task incremental learning benchmarks: CIFAR10-5, CIFAR100-10, CIFAR100-20 and MiniImageNet-20.
CIFAR10-5 is constructed by dividing CIFAR10 \cite{krizhevskyLearningMultipleLayers2009} into 5 tasks and each task has 2 classes.
Similarly, CIFAR100-10 and CIFAR100-20 are constructed by dividing CIFAR100 \cite{krizhevskyLearningMultipleLayers2009} into 10 10-classification tasks and 20 5-classification tasks respectively.
MiniImageNet-20 is constructed by dividing MiniImageNet \cite{vinyalsMatchingNetworksOne2016} into 20 tasks and each task has 5 classes.
We leave the results on the CIFAR10-5 and CIFAR100-20 in the supplementary.
\paragraph{Baselines}
% \subsubsection{Baselines}
%tab:exp_performance_size
\begin{table}[t]
\centering
\small
\begin{tabular}{ccccc}
\toprule
\multirow{2}{*}{Method}& \multicolumn{2}{c}{CIFAR100-10 (10 tasks)}& \multicolumn{2}{c}{$\mathcal{S}^\text{long}$ (100 tasks)} 
\\
& AP& $\mathcal{M}$(M)& AP& $\mathcal{M}$(M)\\
\midrule
Finetune& 0.180& 0.607& 0.200& 0.607\\
Independent& 0.820& 6.070& 0.570& 60.705\\
PN&  0.821& 12.422& -& -\\
Learn to Grow& 0.791& 1.926& -& -\\
MNTDP-S& 0.750& 6.100& 0.680& 39.658\\
MNTDP-D& 0.830& 6.075& 0.750& 25.508\\
\midrule
PAR& \textbf{0.853}& 1.400& \textbf{0.773}& 13.475\\
\bottomrule
\end{tabular}
\caption{Compared with existing parameter allocation methods based on dynamic model, PAR obtains better average performance (AP) with fewer model parameters ($\mathcal{M}$, M=1e6) .
}
\label{tab:exp_size}
\end{table}

%tab:ablation
\begin{table}[t]
\centering
\small
\begin{tabular}{cccccccc}
\toprule
\multirow{2}{*}{\#}&
\multicolumn{3}{c}{Allocation Strategy}&
\multirow{2}{*}{Reg.}&
\multirow{2}{*}{AP}&
% \multirow{2}{*}{AF}&
\multirow{2}{*}{$\mathcal{M}$(M)}&
\multirow{2}{*}{T(h)}\\
&
Fixed&
% Fine-grained Search&
FS&
% Coarse-grained Search&
CS& & & & \\
\midrule
1& \checkmark& & & & 0.855&
% 0.000&
2.236& 1.0\\
2& & \checkmark& & & 0.861&
% 0.000&
2.277& 1.9\\
3& & \checkmark& \checkmark& & 0.870&
% 0.000&
1.900& 1.2\\
4& & & & \checkmark& 0.775&
% -0.091&
0.244& 1.0\\
5& \checkmark& & & \checkmark& 0.834&
% -0.027&
1.129&
1.0\\
6& & \checkmark& & \checkmark& 0.856&
% -0.013&
1.210&
1.4\\
\midrule
PAR& & \checkmark& \checkmark& \checkmark& 0.853&
% -0.020&
1.400& 1.1\\
\bottomrule
\end{tabular}
\caption{Ablation studies of components in PAR on the CIFAR100-10.
The Fixed represents using a fixed cell from DARTs.
The FS and CS represent fine-grained and coarse-grained search space respectively.
The Reg. represents the parameter regularization.
The PAR makes the trade off among average performance (AP), parameter overhead ($\mathcal{M}$), and time overhead (T). 
}
\label{tab:exp_ablation}
\end{table}

\begin{figure*}[t]
    \centering
    % \small
    \begin{subfigure}{0.23\linewidth}
    \includegraphics[width=1\linewidth]{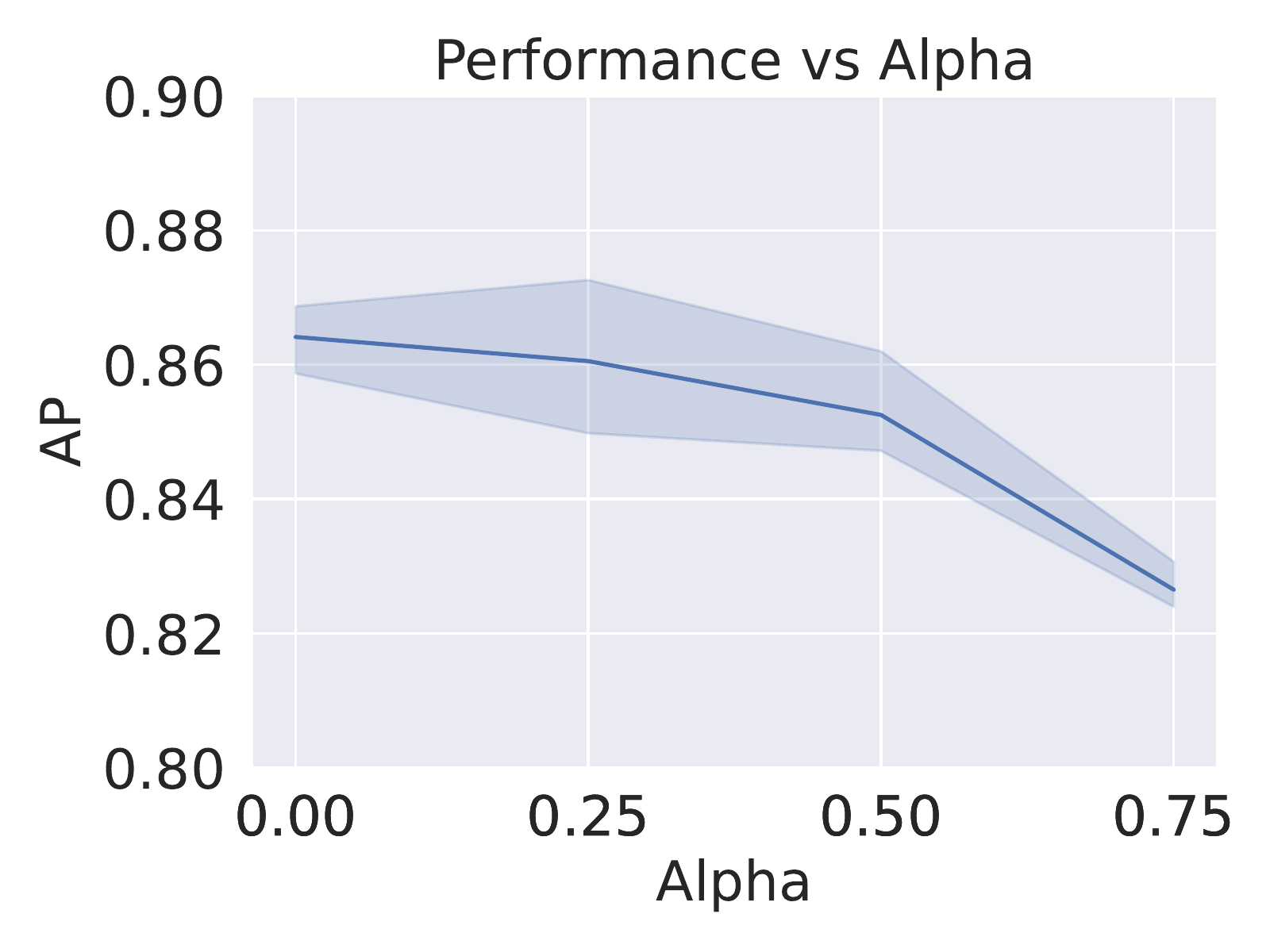}
    \caption{Average performance (AP)}
    \label{fig:performace_alpha}
    \end{subfigure}
    \hfill
    \begin{subfigure}{0.23\linewidth}
    \includegraphics[width=1\linewidth]{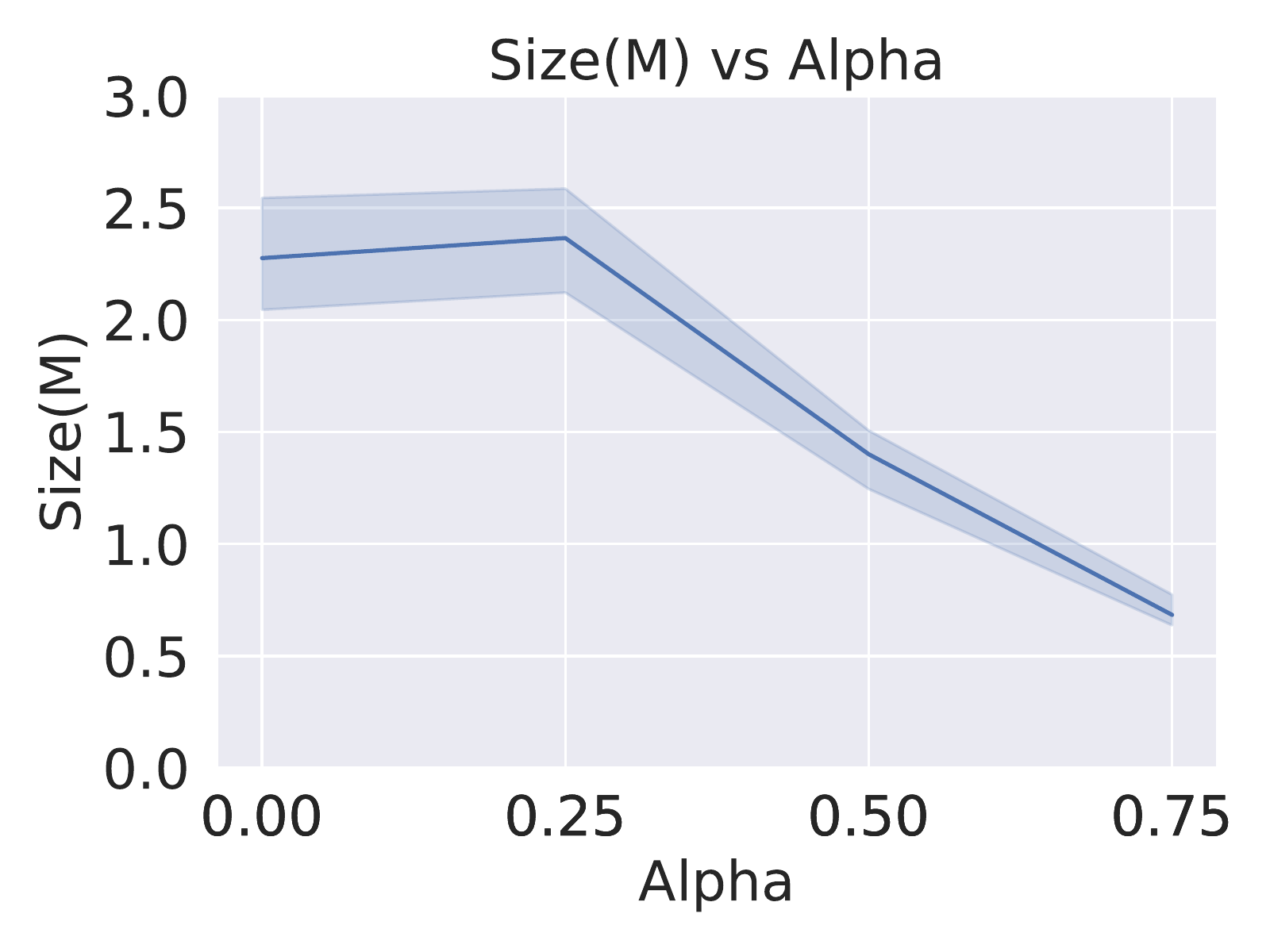}
    \caption{Model size}
    \label{fig:size_alpha}
    \end{subfigure}
    \hfill
    \begin{subfigure}{0.23\linewidth}
    \includegraphics[width=1\linewidth]{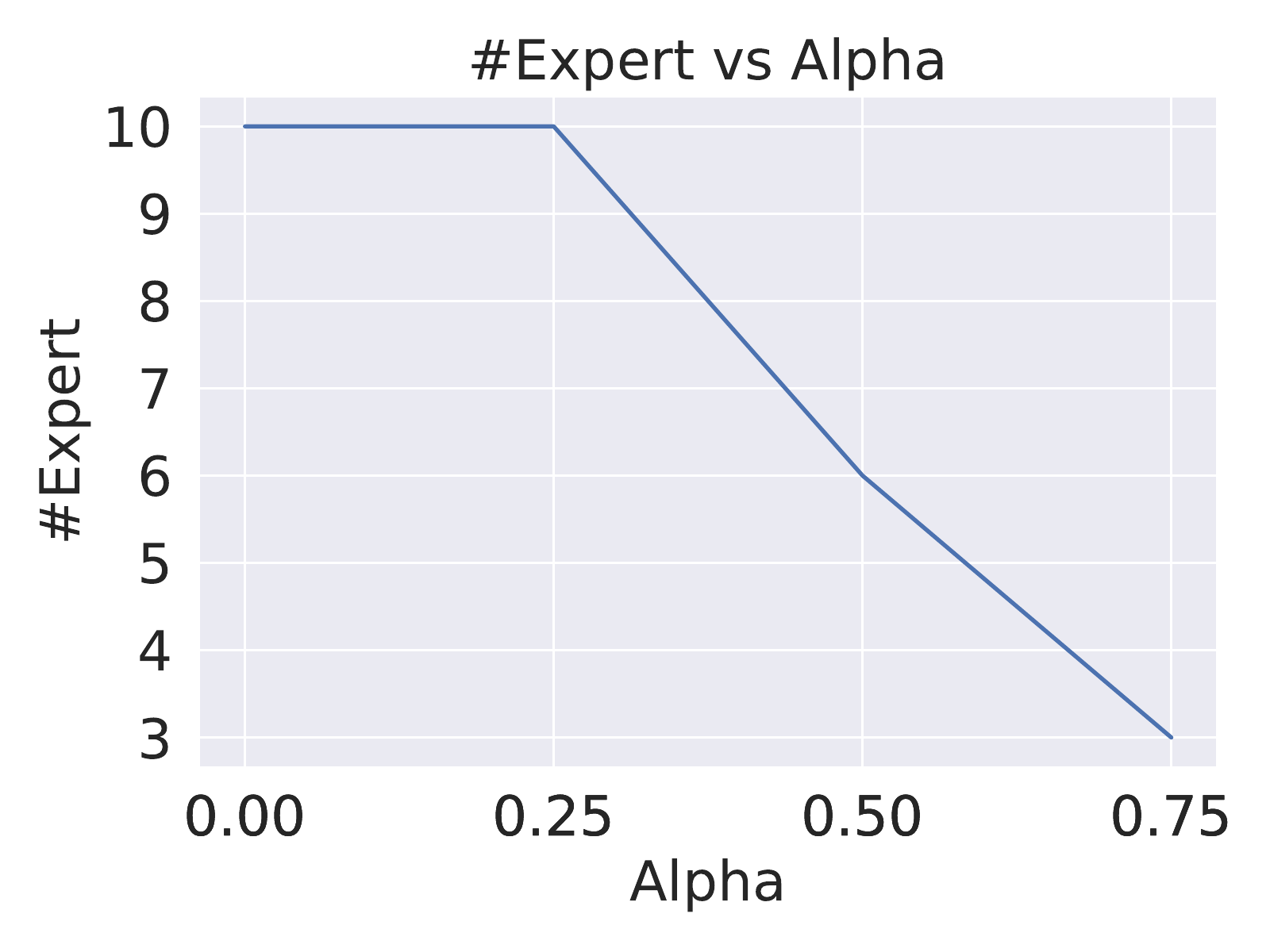}
    \caption{Experts}
    \label{fig:experts_alpha}
    \end{subfigure}
    \hfill
    \begin{subfigure}{0.23\linewidth}
    \includegraphics[width=1\linewidth]{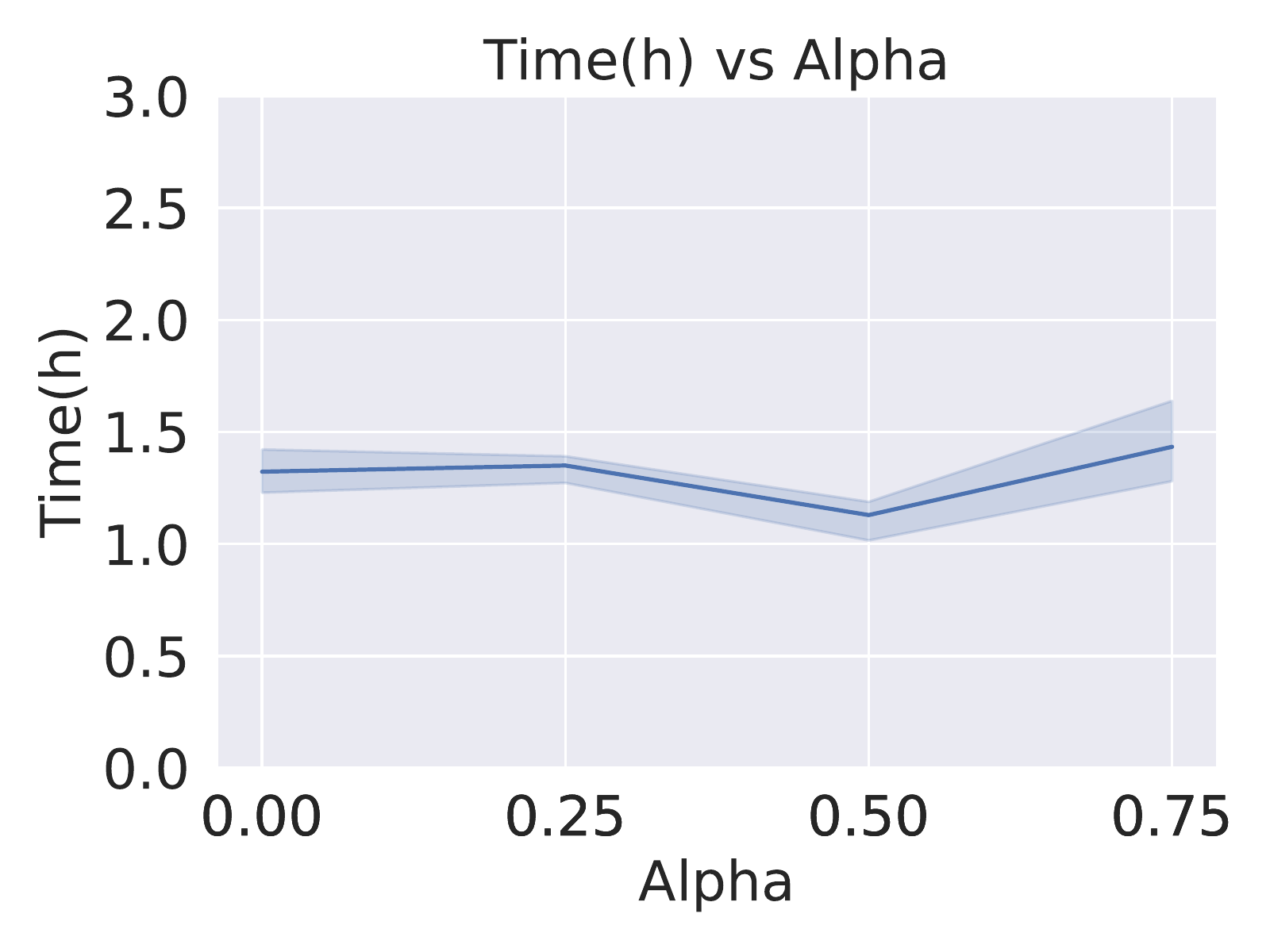}
    \caption{Time}
    \label{fig:time_alpha}
    \end{subfigure}
    \caption{
    Ablation studies of $\alpha$ (Alpha) when $\beta=0.75$.
    The larger $\alpha$ encourages the model to reuse experts. so the model size and number of experts decreases with the increase of $\alpha$.
    The $\alpha$ is independent of the architecture search, so the learning time is stable relative to it.
    }
    \label{fig:exp_alpha}
\end{figure*}

\begin{figure*}[t]
    \centering
    % \small
    \begin{subfigure}{0.23\linewidth}
    \includegraphics[width=1\linewidth]{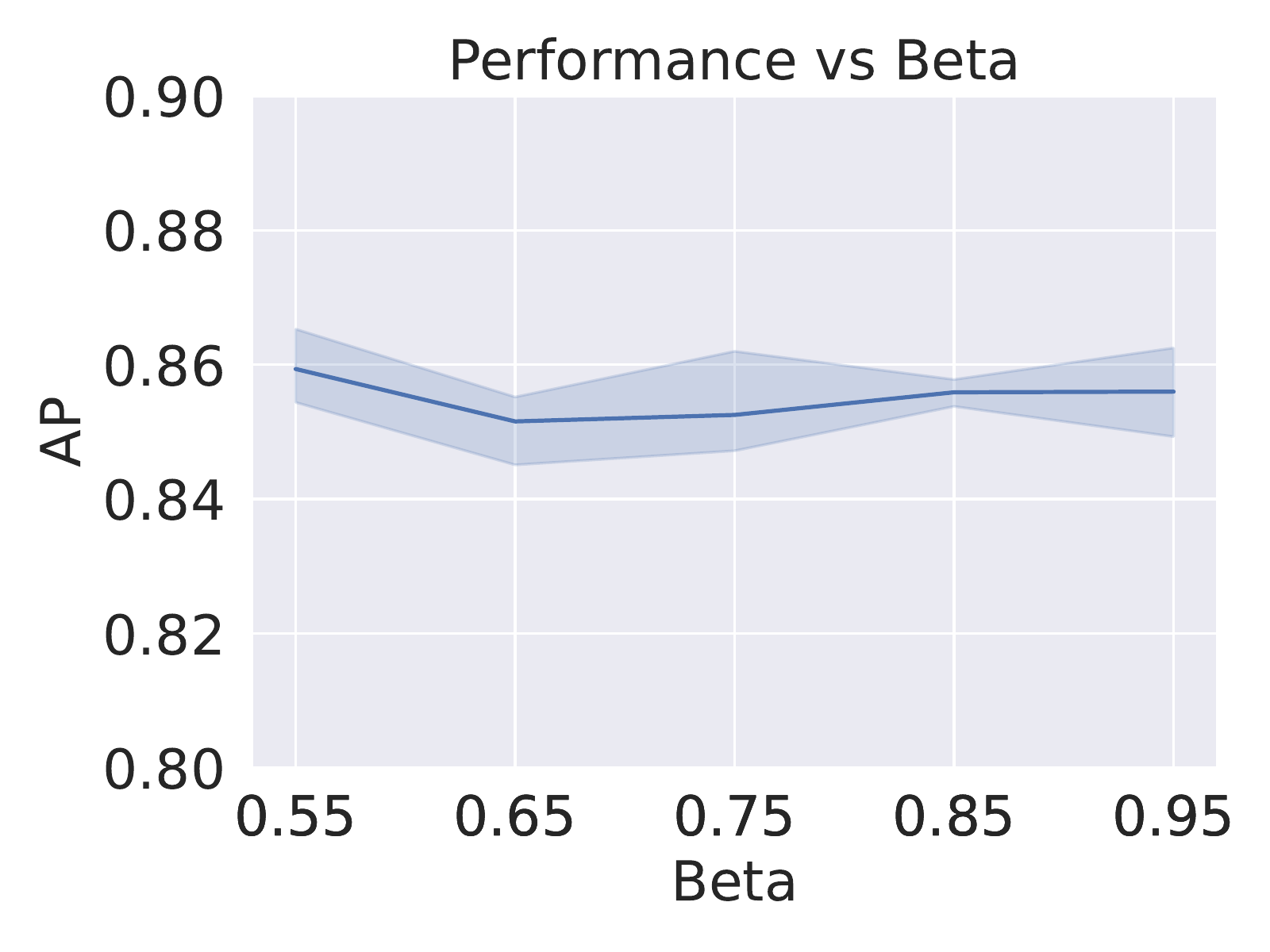}
    \caption{Average performance (AP)}
    \label{fig:performace_beta}
    \end{subfigure}
    \hfill
    \begin{subfigure}{0.23\linewidth}
    \includegraphics[width=1\linewidth]{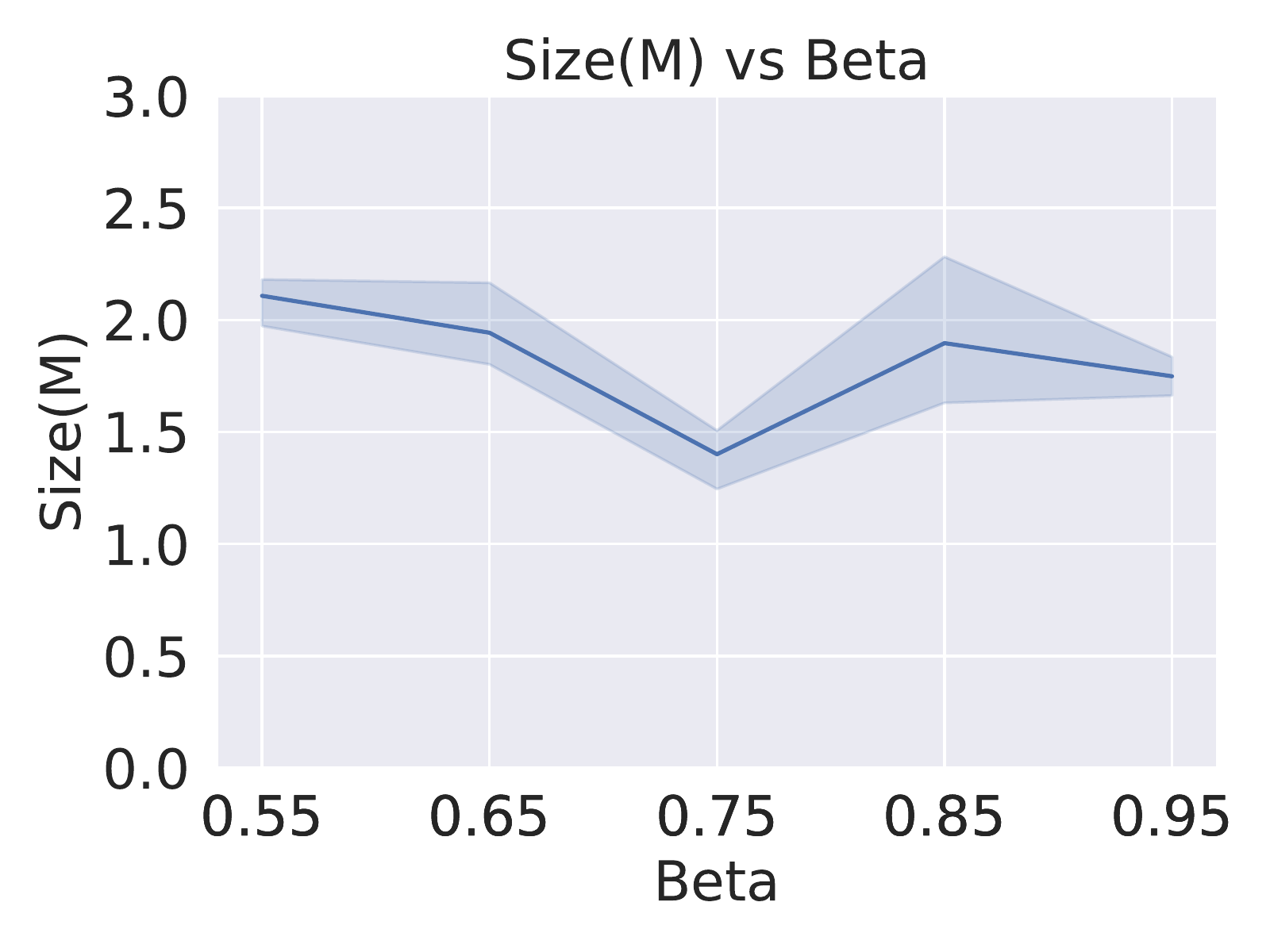}
    \caption{Model size}
    \label{fig:size_beta}
    \end{subfigure}
    \hfill
    \begin{subfigure}{0.23\linewidth}
    \includegraphics[width=1\linewidth]{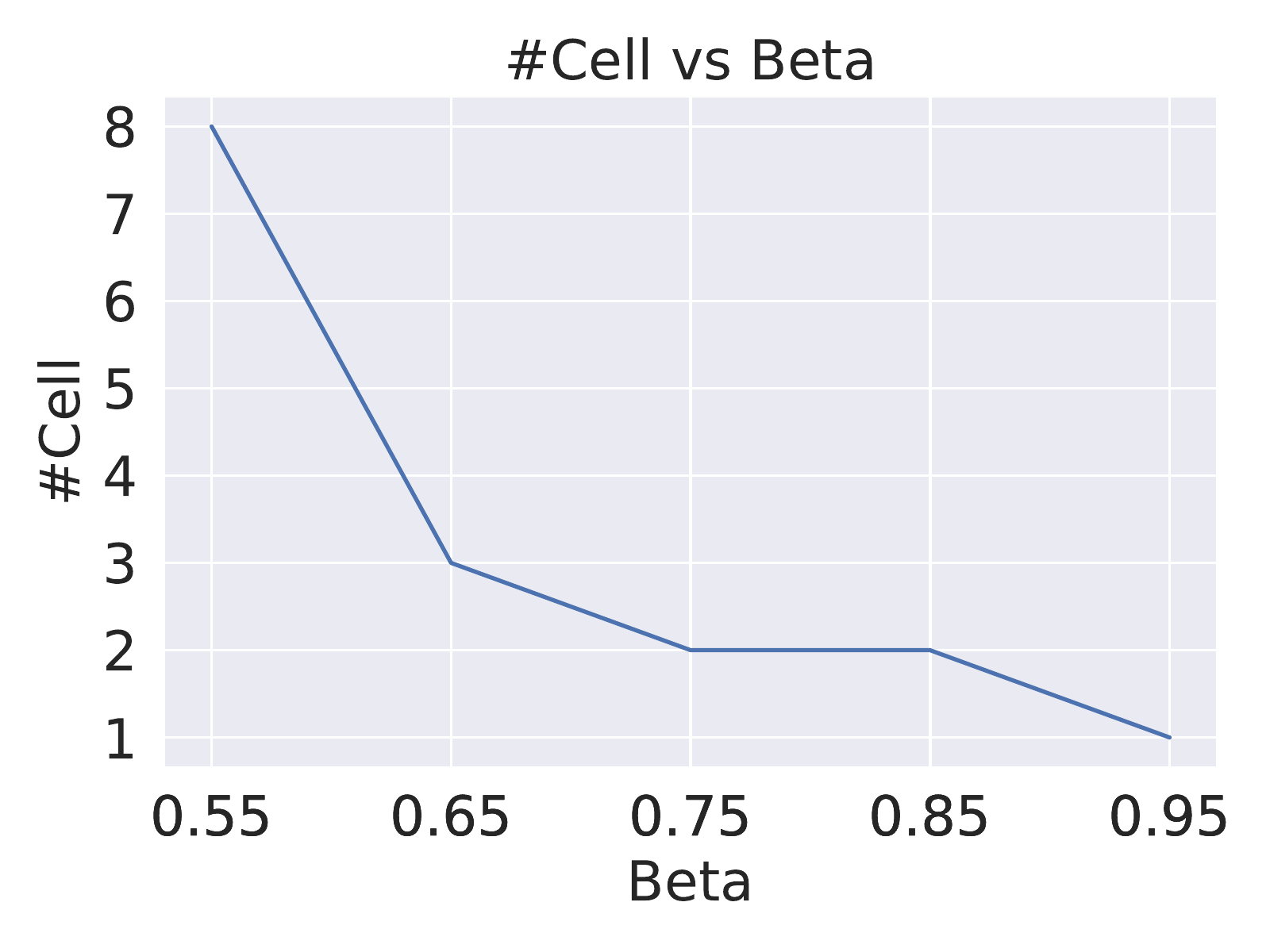}
    \caption{Cells}
    \label{fig:size_beta}
    \end{subfigure}
    \hfill
    \begin{subfigure}{0.23\linewidth}
    \includegraphics[width=1\linewidth]{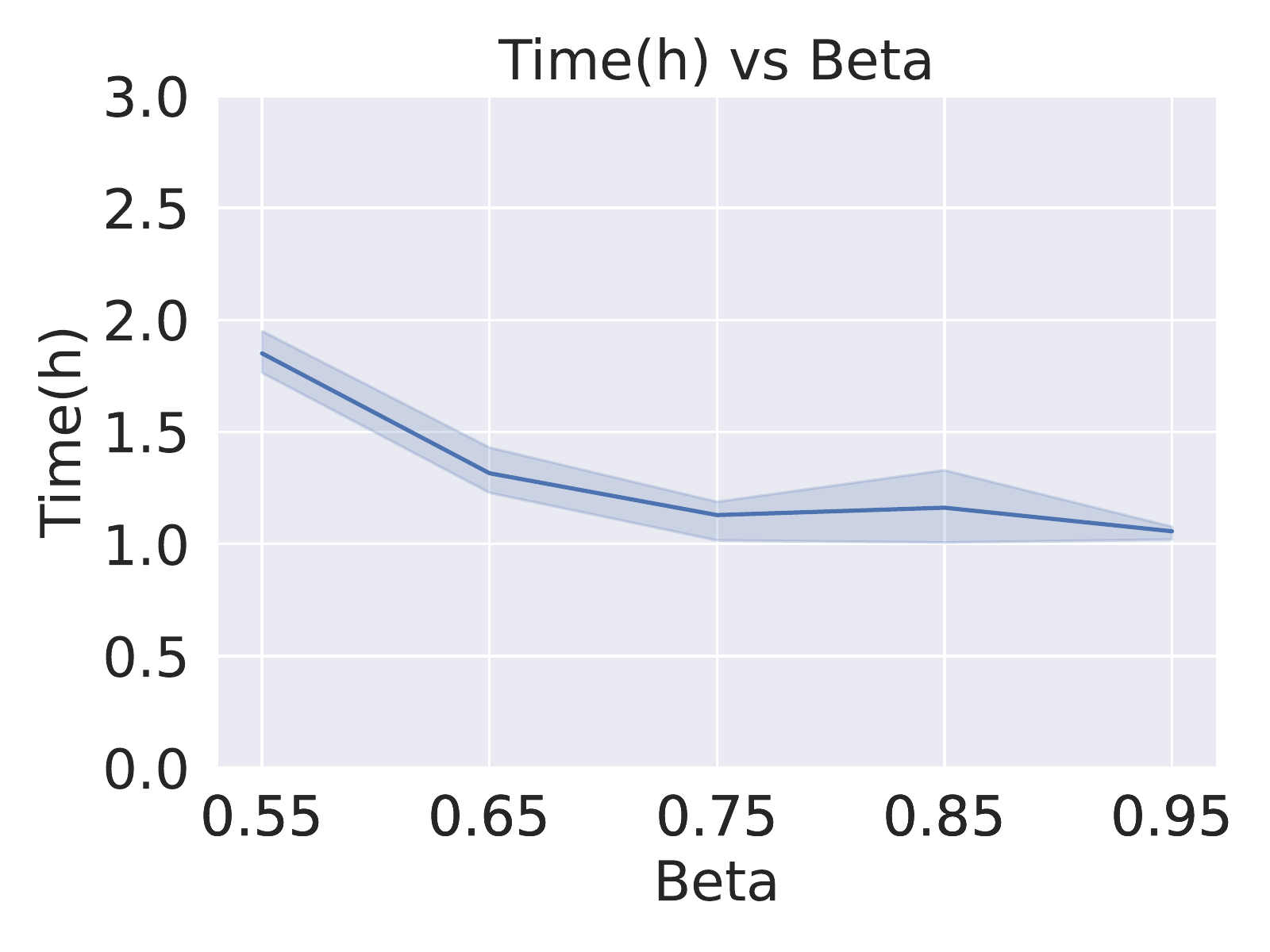}
    \caption{Time}
    \label{fig:time_beta}
    \end{subfigure}
    \caption{
    Ablation studies of $\beta$ (Beta) when $\alpha=0.5$.
    The $\beta$ has little effect on the average performance and size of model, and conversely has a great effect on the learning time because the determine the frequency of architecture search.
    The larger $\beta$ encourages the model to reuse architectures of existing experts.
    }
    \label{fig:exp_beta}
\end{figure*}
Firstly, we compare our method with two simple baselines Finetune and Independent.
While Finetune learns tasks one-by-one without any constraints, Independent builds a model for each task independently.
Then, we present parameter regularization methods, including EWC \cite{kirkpatrickEWC2017}, LwF \cite{liLearningForgetting2017}, and MAS \cite{aljundiMemoryAwareSynapses2018}, and replay methods, including ER \cite{chaudhryTinyEpisodicMemories2019}, GPM \cite{sahaGradientProjectionMemory2020}, and A-GEM~\cite{chaudhryEfficientLifelongLearning2018}. Meanwhile, we compare parameter allocation methods with the static model, including HAT \cite{serraOvercomingCatastrophicForgetting2018} and CAT \cite{keContinualLearningMixed2020}, and parameter allocation methods with the dynamic model, including PN \cite{rusuProgressiveNeuralNetworks2016}, Learn to Grow \cite{liLearnGrowContinual2019}, SG-F \cite{mendezLifelongLearningCompositional2020}, MNTDP \cite{veniatEfficientContinualLearning2020a}, LMC \cite{ostapenkoContinualLearningLocal2021} with PAR.
% We compare our method with several simple baselines, including Finetune that learns tasks one-by-one without any constraints and Independent that builds a model for each task independently. 
% parameter regularization methods including EWC \cite{kirkpatrickEWC2017}, LwF \cite{liLearningForgetting2017}, IMM \cite{leeOvercomingCatastrophicForgetting2017a}, and MAS \cite{aljundiMemoryAwareSynapses2018}; memory based methods including iCaRL \cite{rebuffiICaRLIncrementalClassifier2017}, ER \cite{chaudhryTinyEpisodicMemories2019}, GCL \cite{tangGraphBasedContinualLearning2020}, GPM \cite{sahaGradientProjectionMemory2020} and ACL \cite{ebrahimiAdversarialContinualLearning2020a}; 
% parameter allocation method with static model: HAT \cite{serraOvercomingCatastrophicForgetting2018}, RPSnet \cite{rajasegaranRandomPathSelection2019}, InstAParam \cite{chenMitigatingForgettingOnline2020}, and CAT \cite{keContinualLearningMixed2020}; parameter allocation method with dynamic model: PN \cite{rusuProgressiveNeuralNetworks2016}, Learn to Grow \cite{liLearnGrowContinual2019}, SG-F \cite{mendezLifelongLearningCompositional2020}, MNTDP \cite{veniatEfficientContinualLearning2020a}, LMC \cite{ostapenkoContinualLearningLocal2021}, BNS \cite{qinBNSBuildingNetwork2021}, and FAS \cite{miaoContinualLearningFilter2021}.

% \subsubsection{Metrics}
\paragraph{Metrics}
Denote the performance of the model on $T_j$ after learning task $T_i$
as $r_{i,j}$ where $j\le i$.
Suppose the current task is $T_t$, the \emph{average performance (AP)} and \emph{average forgetting (AF)} are as follows:
\begin{equation}
    AP=\frac{1}{t}\sum_{j=1}^tr_{t,j}, AF=\frac{1}{t}\sum_{j=1}^tr_{t,j}-r_{j,j}.
\end{equation}
To evaluate the parameter overhead, we denote the total number of the model as $\mathcal{M}$.

% \subsubsection{Implementation details}
\paragraph{Implementation details}
We implement PAR by the PyTorch and open the source code\footnote{https://github.com/WenjinW/PAR}.
We set the number of cells of each expert in PAR as 4, and set the $\alpha$ to 0.5 and the $\beta$ to 0.75 for all tasks.
We adopt the SGD optimizer whose initial learning rate is $0.01$ and anneals following a cosine schedule.
We also set the momentum of SGD to $0.9$ and search weight decay from $[0.0003, 0.001, 0.003, 0.01]$ according to the validation performance.
% All results are averaged across three seeds.
The results are averaged across three runs with different random seeds.

Without special instructions, following MNTDP\cite{veniatEfficientContinualLearning2020a}, we adopt a light version of ResNet18 \cite{heDeepResidualLearning2016} with multiple output heads as the backbone for baselines.
For the performance on CTrL, we report the results from MNTDP \cite{veniatEfficientContinualLearning2020a} and LMC \cite{ostapenkoContinualLearningLocal2021}.
For the performance on mixed CIFAR100 and F-CelebA, we report the results from CAT \cite{keContinualLearningMixed2020}.
For the performance on classical benchmarks, we: report the results of RPSnet \cite{rajasegaranRandomPathSelection2019}, InstAParam \cite{chenMitigatingForgettingOnline2020}, and BNS \cite{qinBNSBuildingNetwork2021} from origin papers; adopt the implementation of \cite{buzzegaDarkExperienceGeneral2020} for memory-based methods; and adopt our implementation for the others.

%fig:exp_coarse_cifar100
\begin{figure}[t]
% \small
\centering
\includegraphics[width=1\columnwidth]{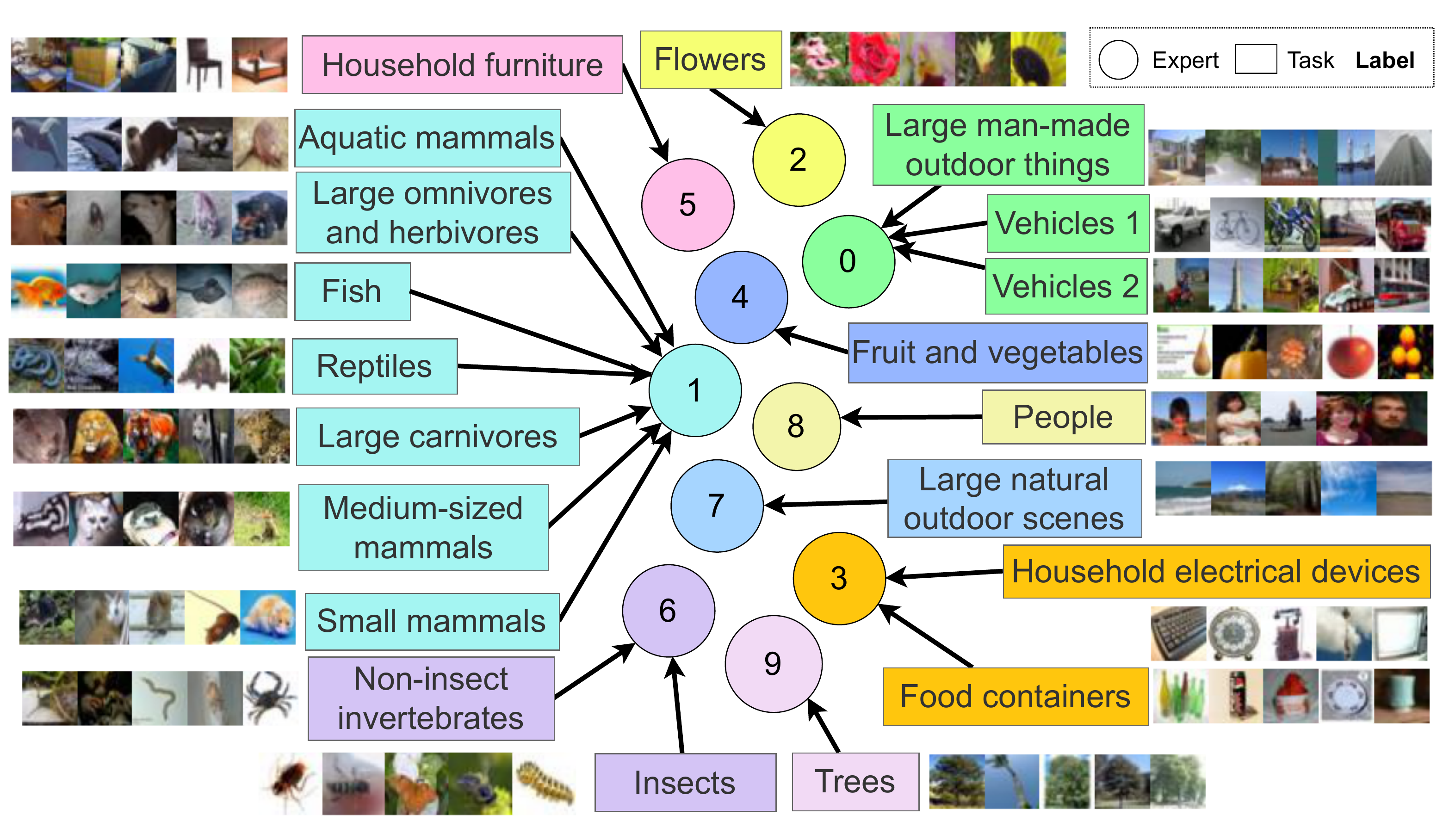}
\caption{Visualization of \emph{task groups} and \emph{experts} on CIFAR100-Coarse.
The PAR groups semantically related tasks into the same group.
We distinguish the different groups by different colors.
}
\label{fig:exp_coarse_cifar100}
\end{figure}

\begin{figure}[t]
    \centering
    % \small
    \begin{subfigure}{0.47\columnwidth}
    \includegraphics[width=1\linewidth]{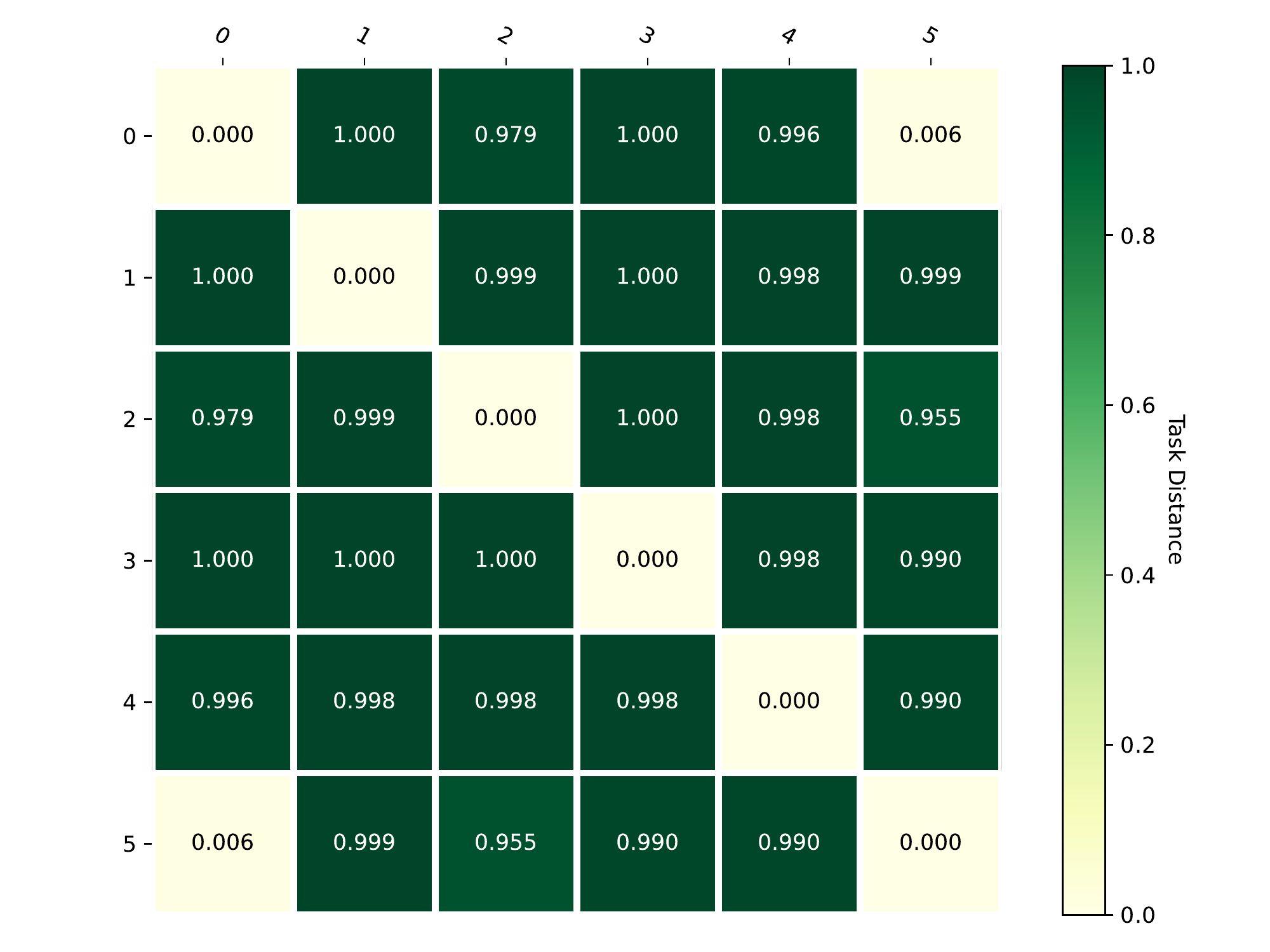}
    \caption{$\mathcal{S}^{-}$}
    \label{fig:task_distance_ctrl_minus}
    \end{subfigure}
    \hfill
    \begin{subfigure}{0.47\columnwidth}
    \includegraphics[width=1\linewidth]{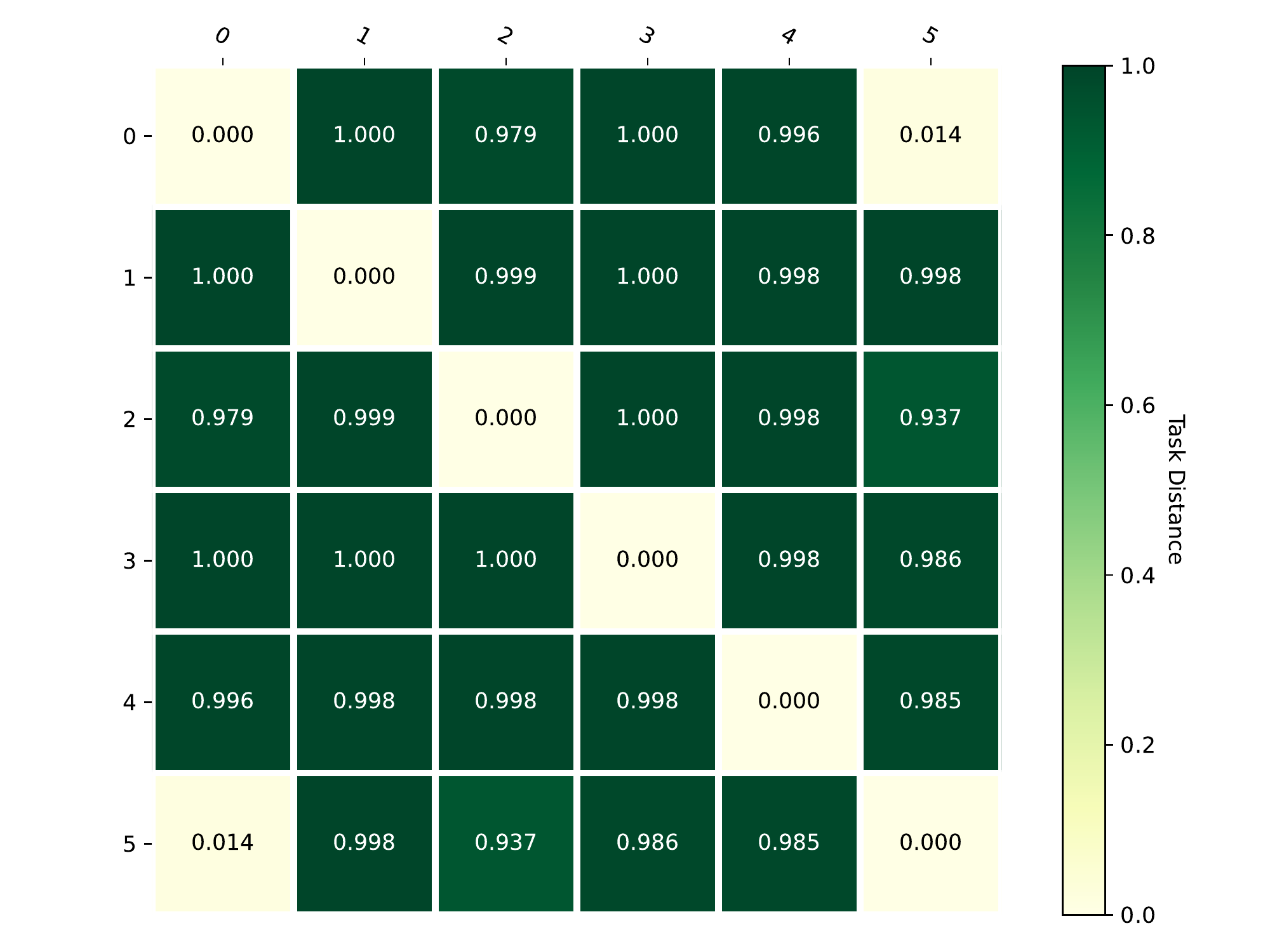}
    \caption{$\mathcal{S}^{+}$}
    \label{fig:task_distance_ctrl_plus}
    \end{subfigure}
    
    \begin{subfigure}{0.47\columnwidth}
    \includegraphics[width=1\linewidth]{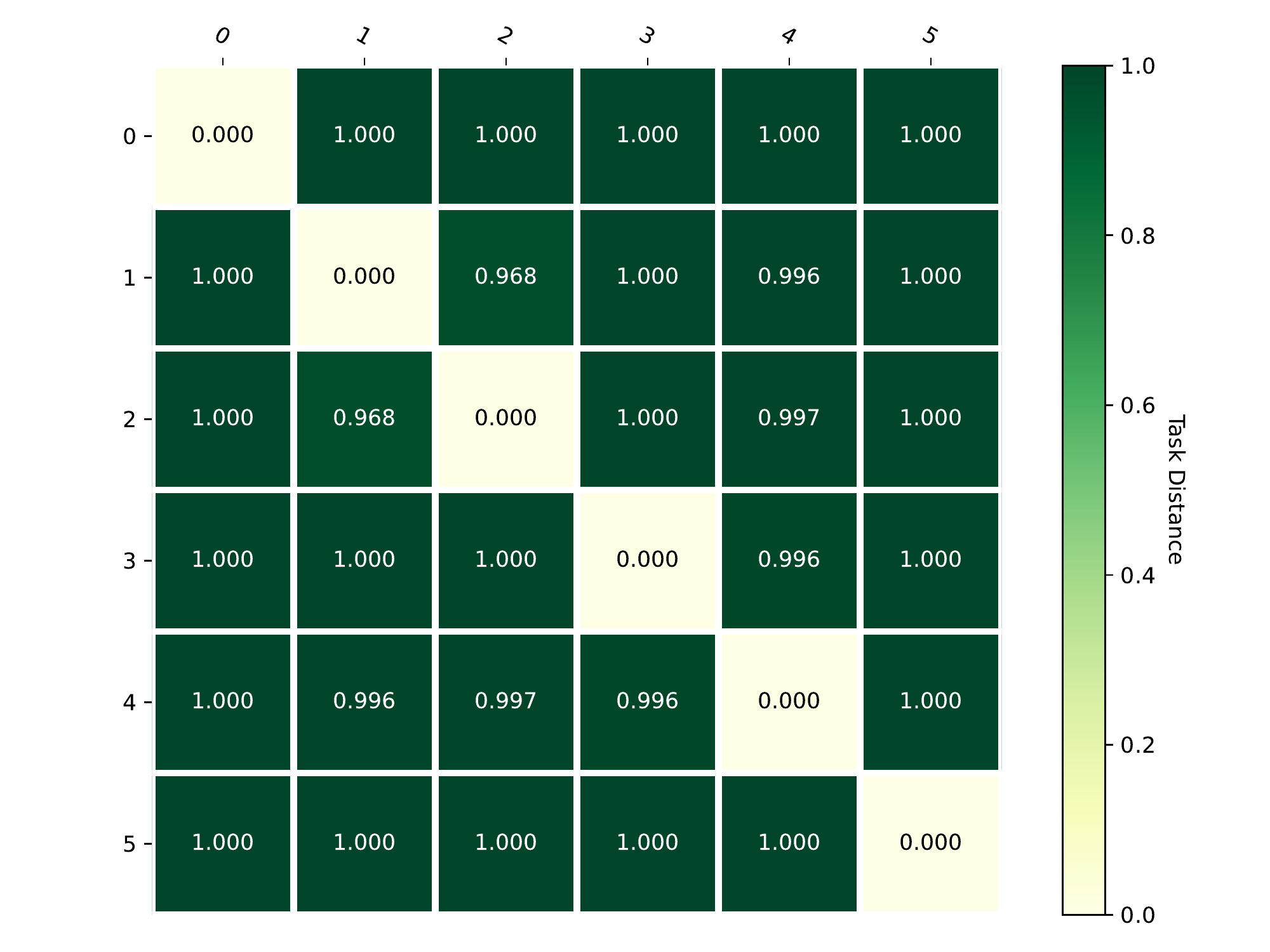}
    \caption{$\mathcal{S}^\text{in}$}
    \label{fig:task_distance_ctrl_in}
    \end{subfigure}
    \hfill
    \begin{subfigure}{0.47\columnwidth}
    \includegraphics[width=1\linewidth]{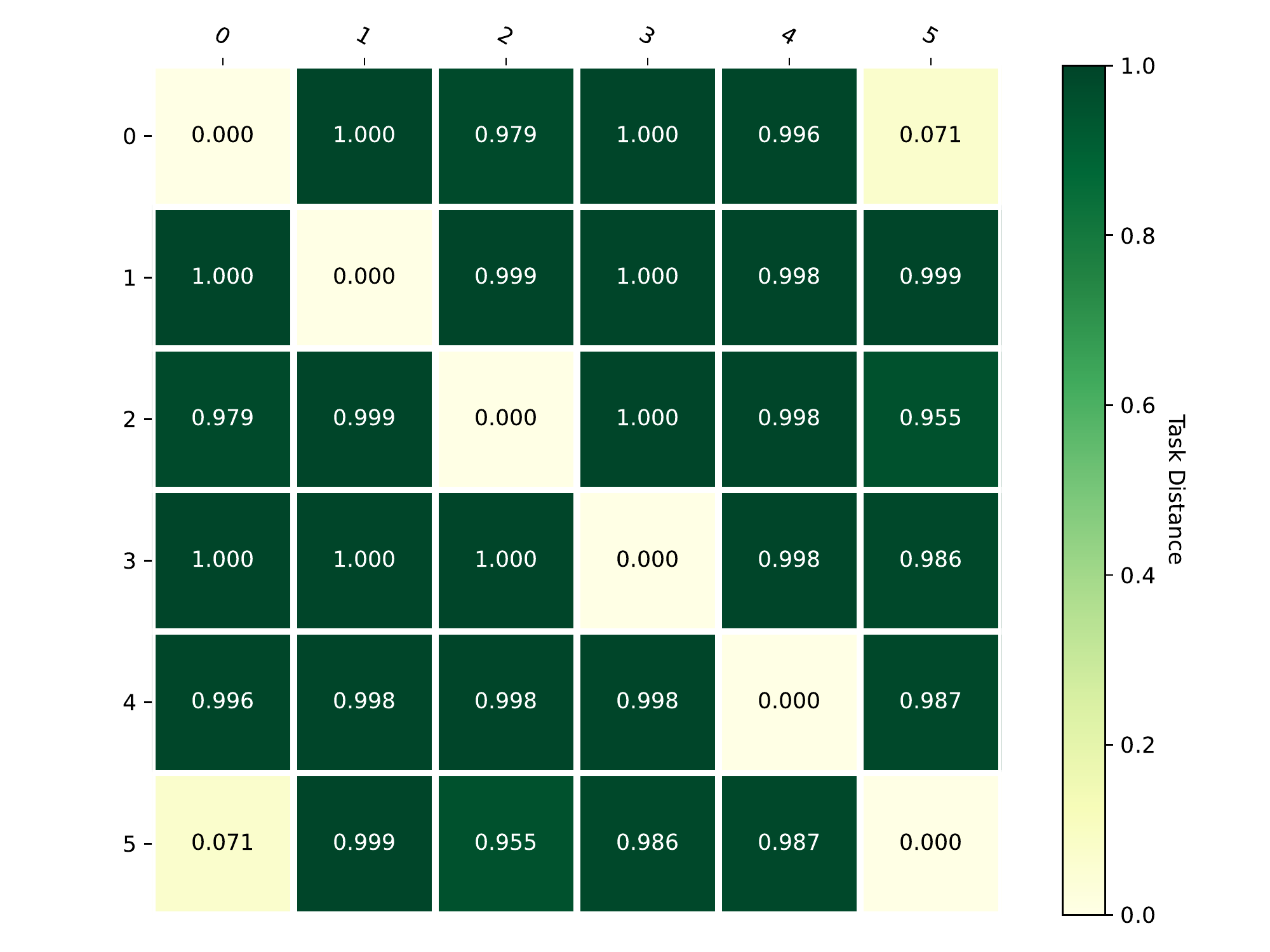}
    \caption{$\mathcal{S}^\text{out}$}
    \label{fig:task_distance_ctrl_out}
    \end{subfigure}
    \caption{
    The heat-maps of task distances of four streams in CTrL benchmark.
    The PAR can recognize the related tasks in $\mathcal{S}^{-}$, $\mathcal{S}^{+}$, and $\mathcal{S}^{\text{out}}$ exactly.
    }
    \label{fig:task_distance_ctrl}
\end{figure}

\subsection{Comparison with baselines}
\label{sec:exp_comparison}

At first, we evaluate PAR on benchmarks with mixed similar and dissimilar tasks.
As shown in \cref{tab:exp_ctrl}, PAR outperforms baselines on all six streams in CTrL.
One reason is that PAR allows knowledge transfer among related tasks while preventing interference among tasks not related.
For example, performance on stream $\mathcal{S}^+$ shows that PAR can update the knowledge of the expert for the task $t_1^-$ by the task $t_1^+$ with the same distribution and more samples.
Performance on streams $\mathcal{S}^-$ and $\mathcal{S}^\text{out}$ shows that our method can transfer knowledge from experts of tasks related to the new task.
Another reason is that a task-tailored architecture helps each expert perform better.
Moreover, performance on stream $\mathcal{S}^\text{long}$ shows that PAR is scalable for the long task sequence.
Similarly, results in \cref{tab:exp_cifar100_celeba} show that PAR outperforms parameter allocation methods with static models.

Further, we evaluate the PAR on classical benchmarks. Results in \cref{tab:exp_cifar100-10_cifar10} show that PAR outperforms parameter regularization methods, memory-based methods, and parameter allocation methods with dynamic model on the CIFAR100-10 and MiniImageNet-20.

At last, we analyze the parameter overhead\footnote{We ignore the parameter overhead of the extractor (11.18M) in divergence estimation, which is fixed and will not become a bottleneck with the increase of tasks.} of PAR on CIFAR100-10 and $\mathcal{S}^\text{long}$.
Compared with baselines, the PAR obtains better average performance with fewer model parameters.
The reason is that the the allocated experts in PAR have compact architecture and are parameter efficient.
Performance on the stream $\mathcal{S}^\text{long}$ shows that PAR is scalable.

\subsection{Ablation Studies}
\label{sec:exp_ablation}
We analyze the effects of components in PAR and the results are listed in \cref{tab:exp_ablation}.
First, we evaluate the impact of hierarchical architecture search on parameter allocation. 
Compared with using a fixed cell from DARTs \cite{liuDARTSDIFFERENTIABLEARCHITECTURE2018} (\#1), searching cells from fine-grained space can improve model performance (\#2).
Combined with coarse-grained space, searching from the hierarchical space can further improve performance while reducing the time and parameter overhead (\#3).
Then, we find that parameter regularization alone is time efficient, but its performance is low (\#4).
Finally, by selecting an appropriate strategy from parameter allocation and regularization based on the learning difficulty of each new task, PAR make a trade off among average performance, parameter overhead, and time overhead.

We analyze the impact of important hyper-parameters $\alpha$ and $\beta$ in PAR on the CIFAR100-10.
The $\alpha$ determines the reuse experts during the learning and the larger $\alpha$ encourages the model to reuse experts.
Results in \cref{fig:exp_alpha} show that the model size and number of experts decreases with the increase of $\alpha$.
The $\beta$ determines the frequency of architecture search and the larger $\beta$ encourages the model to reuse architectures of existing experts 
Results in \cref{fig:exp_beta} show that the learning time decreases with the increase of $\beta$.

\subsection{Visualization}
\label{sec:exp_relatedness}
To analyze the task distance in PAR, we construct a new benchmark CIFAR100-coarse by dividing CIFAR100 into 20 tasks based on the coarse-grained labels.
The illustration in \cref{fig:exp_coarse_cifar100} shows that the PAR can obtain reasonable task groups.
For example, the PAR adaptively puts tasks about animals into the same groups, such as the Aquatic mammals, Large omnivores and herbivores, Fish, and so on.
The Food containers and Household electrical devices are divided into the same group since they both contain cylindrical and circular objects.
It also finds the relation between Non-insect invertebrates and Insects.

Moreover, we visualize the heat maps of task distances obtained by PAR on four streams of CTrL in \cref{fig:task_distance_ctrl}.
The distance between the first and the last task on streams $\mathcal{S}^-$, $S^\mathcal{+}$ and $\mathcal{S}^\text{out}$ are small, which is consistent with the fact that they all come from the same data distribution.
Instead, the distribution of the last task on stream $S^\mathcal{\text{in}}$ is perturbed from the distribution of the first task, so their distance is large.
The above results show that PAR can obtain reasonable task distance and relatedness.

% \section{Limitations}
% \label{sec:limitations}
% \textbf{Pre-trained extractor in distance calculation}

% \textbf{Architecture search}
% Despite

\section{Conclusion}
In this paper, we propose a new method named \textbf{P}arameter \textbf{A}llocation \& \textbf{R}egularization (PAR).
It allows the model to adaptively select an appropriate strategy from parameter allocation and regularization for each task based on the task learning difficulty.
% To measure the learning difficulty according to relevance among tasks, we propose a divergence estimation method via Nearest-Prototype distance to calculate the distance between tasks.
Experimental results on multiple benchmarks demonstrate that PAR is scalable and significantly reduces the model redundancy while improving performance.
Moreover, qualitative analysis shows that PAR can obtain reasonable task relatedness.
Our method is flexible, and in the future, we will introduce more relatedness estimation, regularization, and allocation strategies into PAR.

\section*{Acknowledgments}
This work was supported by the NSFC projects (No.~62072399, No.~U19B2042), Chinese Knowledge Center for Engineering Sciences and Technology, MoE Engineering Research Center of Digital Library, China Research Centre on Data and Knowledge for Engineering Sciences and Technology, and the Fundamental Research Funds for the Central Universities.

\newpage
%%%%%%%%% REFERENCES
{\small
\bibliographystyle{ieee_fullname}
\bibliography{ref}
}
\end{document}